\documentclass[11pt]{article}

\usepackage[final]{acl}
\usepackage{amsmath}
\usepackage{amssymb}
\usepackage{booktabs}
\usepackage{multirow}
\usepackage{svg}
\usepackage{times}
\usepackage{latexsym}
\usepackage[T1]{fontenc}

\usepackage[utf8]{inputenc}

\usepackage{microtype}
\usepackage{enumitem}
\usepackage{inconsolata}

\usepackage{graphicx}
\usepackage{tcolorbox}
\definecolor{promptbg}{gray}{0.95}
\definecolor{promptframe}{gray}{0.6}

\title{Mitigating Action-Relation Hallucinations in LVLMs via Relation-aware Visual Enhancement}

\author{
Zhenxin Qin, Qiang Li, Qingzhuo Wang, Ruiyang Qin, Zhihua Wei, Wen Shen\thanks{Corresponding author.} \\
Tongji University, Shanghai, China \\
\texttt{\{zhenxinqin, qli, 2534123, 2432008, zhihua\_wei, wenshen\}@tongji.edu.cn}
}

\begin{document}
\maketitle
\begin{abstract}
Large Vision-Language Models (LVLMs) have achieved remarkable performance on diverse vision-language tasks. However, LVLMs still suffer from hallucinations, generating text that contradicts the visual input. Existing research has primarily focused on mitigating object hallucinations, but often overlooks more complex relation hallucinations, particularly action relations involving interactions between objects.
In this study, we empirically observe that the primary cause of action-relation hallucinations in LVLMs is the insufficient attention allocated to visual information. Thus, we propose a framework to locate action-relevant image regions and enhance the LVLM's attention to those regions. Specifically, we define the Action-Relation Sensitivity (ARS) score to identify attention heads that are most sensitive to action-relation changes, thereby localizing action-relevant image regions that contain key visual cues. Then, we propose the Relation-aware Visual Enhancement (RVE) method to enhance the LVLM's attention to these action-relevant image regions.
Extensive experiments demonstrate that, compared to existing baselines, our method achieves superior performance in mitigating action-relation hallucinations with negligible additional inference cost. Furthermore, it effectively generalizes to spatial-relation hallucinations and object hallucinations. The code is available at \url{https://github.com/LandHqzx/ARS-RVE}.
\end{abstract}

\section{Introduction}
Large Vision-Language Models (LVLMs) \cite{lu2024deepseek,liu2024survey,chen2024internv} have shown exceptional capabilities across various multimodal applications, ranging from visual question answering to complex reasoning. Despite these advancements, they are still prone to hallucinations, generating answers that are inconsistent with image content \cite{chen2024unified,kaul2024throne,gunjal2024detecting}. This problem reduces the reliability of LVLMs in practical applications.

\begin{figure}[t]
    \centering
    \includegraphics[width=1.0\linewidth]{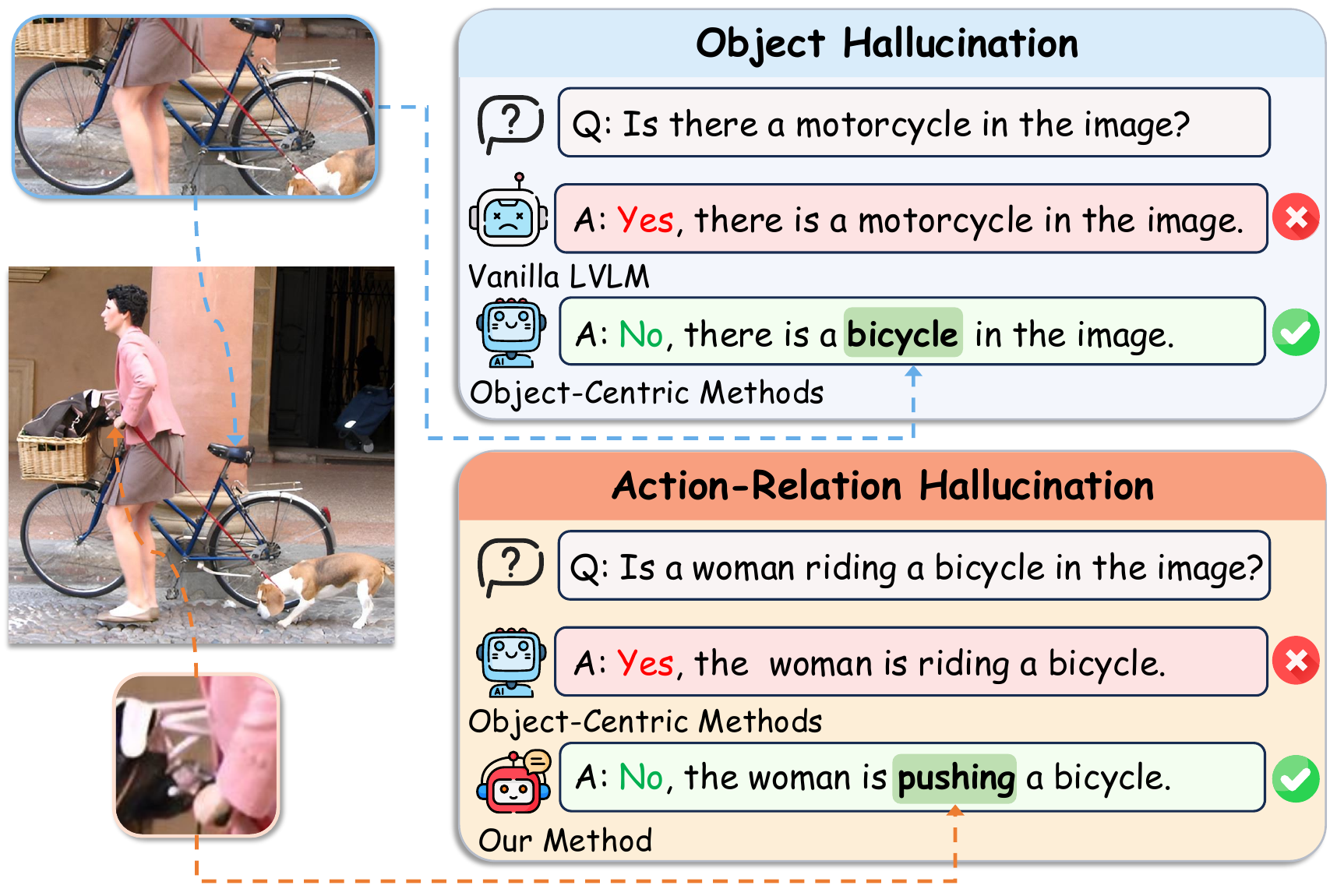}
    \caption{\textbf{Comparison between object hallucination and action-relation hallucination.}
    Existing object-centric methods effectively mitigate object hallucinations (top), but they fail to mitigate action-relation hallucinations (bottom).}
    \label{fig:action_hallucination}
\end{figure}

Numerous methods have been proposed to mitigate hallucinations in LVLMs. Specifically, several studies adopt fine-tuning approaches, such as utilizing preference alignment training \cite{zhao2023beyond,sun2024aligning} and training data refinement \cite{Yu_2024_CVPR}.
To avoid high training costs, other studies propose training-free methods, including  modifying output logits~\cite{chen2024halc,an2025mitigating} and adjusting internal attention \cite{yin2025clearsight,jiang2025devils}. However, these training-free studies are primarily designed for object hallucinations and fail to address action-relation hallucinations. As illustrated in \autoref{fig:action_hallucination}, action-relation hallucinations involve not only objects but also complex interactions between them, requiring LVLMs to precisely capture action-relevant visual features. Please see Appendix~\ref{sec:appendix_gap_analysis} for further comparative analysis.

In this paper, we aim to mitigate relation hallucinations in LVLMs, with a particular focus on action relations. To this end, we first explore the reason behind action-relation hallucinations. We observe that despite image tokens constituting the vast majority  of the input sequence, they receive disproportionately low attention compared to text tokens.
Inspired by the above observation, we propose to enhance the LVLM's attention to action-relevant image tokens to mitigate action-relation hallucinations. Specifically, we propose a framework to first locate action-relevant image regions and  then enhance the LVLM's attention towards these regions during inference.

To locate action-relevant image regions, we define the Action-Relation Sensitivity (ARS) score to measure the sensitivity of each attention head to action-relation changes. Visualization results verify that attention heads with high ARS scores explicitly focus on action-relevant image regions, thereby validating the effectiveness of the ARS score. Notably, we find that the middle layers have higher ARS scores than shallow and deep layers, indicating the middle layers are the most sensitive to action-relation changes.

Therefore, we propose the Relation-aware Visual Enhancement (RVE) method to enhance middle layers' attention toward action-relevant image regions, thereby mitigating action-relation hallucinations. Specifically, for each selected layer, we construct an enhancement mask  based on attention heads with high ARS scores to locate action-relevant image regions. Besides, we observe that attention heads with low ARS scores often capture background noise common to all attention heads. Thus, we construct a denoising mask to locate these action-irrelevant image regions that should not be enhanced.  Finally, we apply these masks to the attention maps of all heads within the selected layer to amplify focus on action-relevant regions.

Our main contributions are as follows:
\begin{itemize}[leftmargin=*]
\item We define the ARS score to quantify the sensitivity of attention heads to action-relation changes. Leveraging this score, we reveal that middle layers are most sensitive to action-relation changes.
\item We propose the training-free RVE method to enhance attention on action-relevant regions while reducing the interference of background noise.
\item Extensive experiments demonstrate that our method can effectively mitigate action-relation hallucinations while generalizing to spatial-relation and object hallucinations, with negligible additional inference cost.
\end{itemize}

\section{Preliminaries and Motivation}
\noindent \textbf{LVLM Generation.} 
Given an input text $T=\{t_1, \dots, t_{N_T}\}$ comprising $N_T$ text tokens and an input image $I=\{v_1, \dots, v_{N_I}\}$ comprising $N_I$ image tokens, an LVLM employs a visual encoder followed by a modality connector to transform the input image tokens into visual embeddings $X_v \in \mathbb{R}^{N_I \times d}$, where $d$ denotes the embedding dimension. For the text, a tokenizer converts the input text into text tokens, followed by an embedding module to obtain text embeddings $X_t \in \mathbb{R}^{N_T \times d}$. Subsequently, the LLM backbone processes the concatenated embeddings of $X_v$ and $X_t$ to autoregressively generate the next tokens.

\noindent \textbf{Multi-Head Attention.}
The LLM backbone relies on the multi-head attention mechanism to integrate visual and textual information from input embeddings. For the $h$-th head (where $h=1, \dots, H$) in the $l$-th layer, we denote the full attention weights of the last token with respect to \textit{all input tokens} as $W^{(l,h)} \in \mathbb{R}^{1 \times N}$, where $N = N_T + N_I$.
We then obtain attention weights corresponding to the \textit{image tokens}, denoted as $A^{(l,h)} \in \mathbb{R}^{1 \times N_I}$, and the corresponding pre-softmax attention scores $S^{(l,h)} \in \mathbb{R}^{1 \times N_I}$.

\begin{figure}[t]
    \centering
    \includegraphics[width=\linewidth]{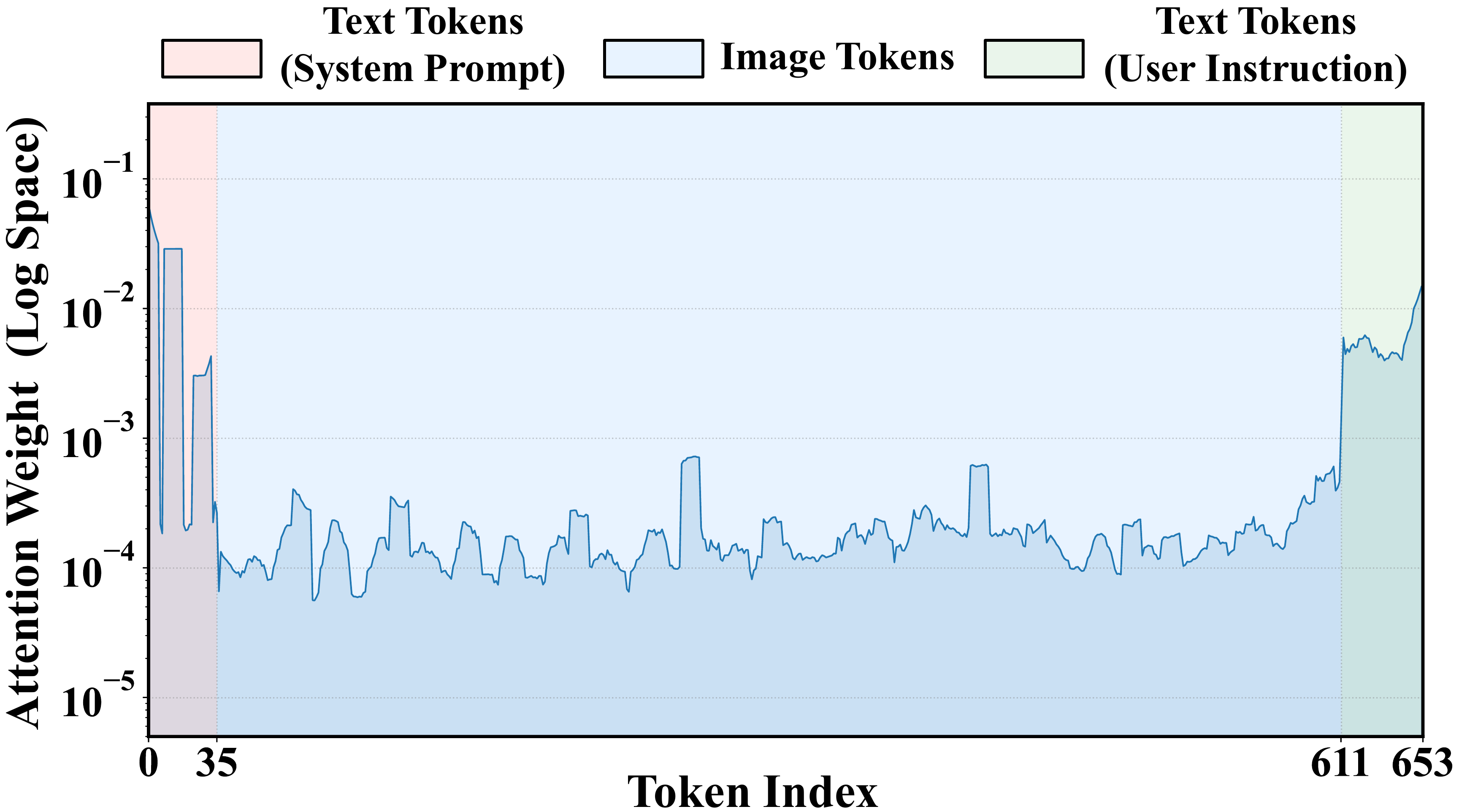} 

    \vspace{-1mm}
    \caption{Attention weights (in the log space) of the last token allocated to the entire input sequence. Results reveal that the attention allocated to text tokens is approximately 10 to 100 times that of image tokens.}
    \label{fig:attn_dist}
\end{figure}

\noindent \textbf{Motivation.} 
To investigate the underlying causes of action-relation hallucinations, we analyze the attention distribution of LLaVA-1.5-7B over the entire input sequence during the generation process. 
As illustrated in Figure~\ref{fig:attn_dist}, our observations reveal a severe modality imbalance: the attention allocated to text tokens is approximately 10 to 100 times that allocated to image tokens.
This suggests that the model may rely heavily on language priors while overlooking the visual information.

To further quantify such imbalance of attention weights, let $\mathcal{T}$ denote the set of text token indices and $\mathcal{I}$ denote the set of image token indices.
We quantify the ratio of attention allocated to text tokens and the ratio for vision tokens as follows:
\begin{equation}
    \begin{aligned}
        r_{\text{att}}^{(l), t} &= \frac{\sum_{i \in \mathcal{T}} W_{1,i}^{(l)}}{\sum_{j=1}^{N} W_{1,j}^{(l)}}, & r_{\text{att}}^{(l), v} &= \frac{\sum_{i \in \mathcal{I}} W_{1,i}^{(l)}}{\sum_{j=1}^{N} W_{1,j}^{(l)}}, 
    \end{aligned}
    \label{eq:ratio_att}
\end{equation}
where $W^{(l)} = \frac{1}{H}\sum_{h=1}^{H} W^{(l,h)}$.
We also calculate the number ratio of text tokens and the number ratio of vision tokens as follows:
\begin{equation}
    \begin{aligned}
        r_{\text{num}}^{t} &= \frac{N_T}{N}, & r_{\text{num}}^{v} &= \frac{N_I}{N}.
    \end{aligned}
    \label{eq:ratio_num}
\end{equation}

\begin{figure}[t]
    \centering
    \includegraphics[width=\linewidth]{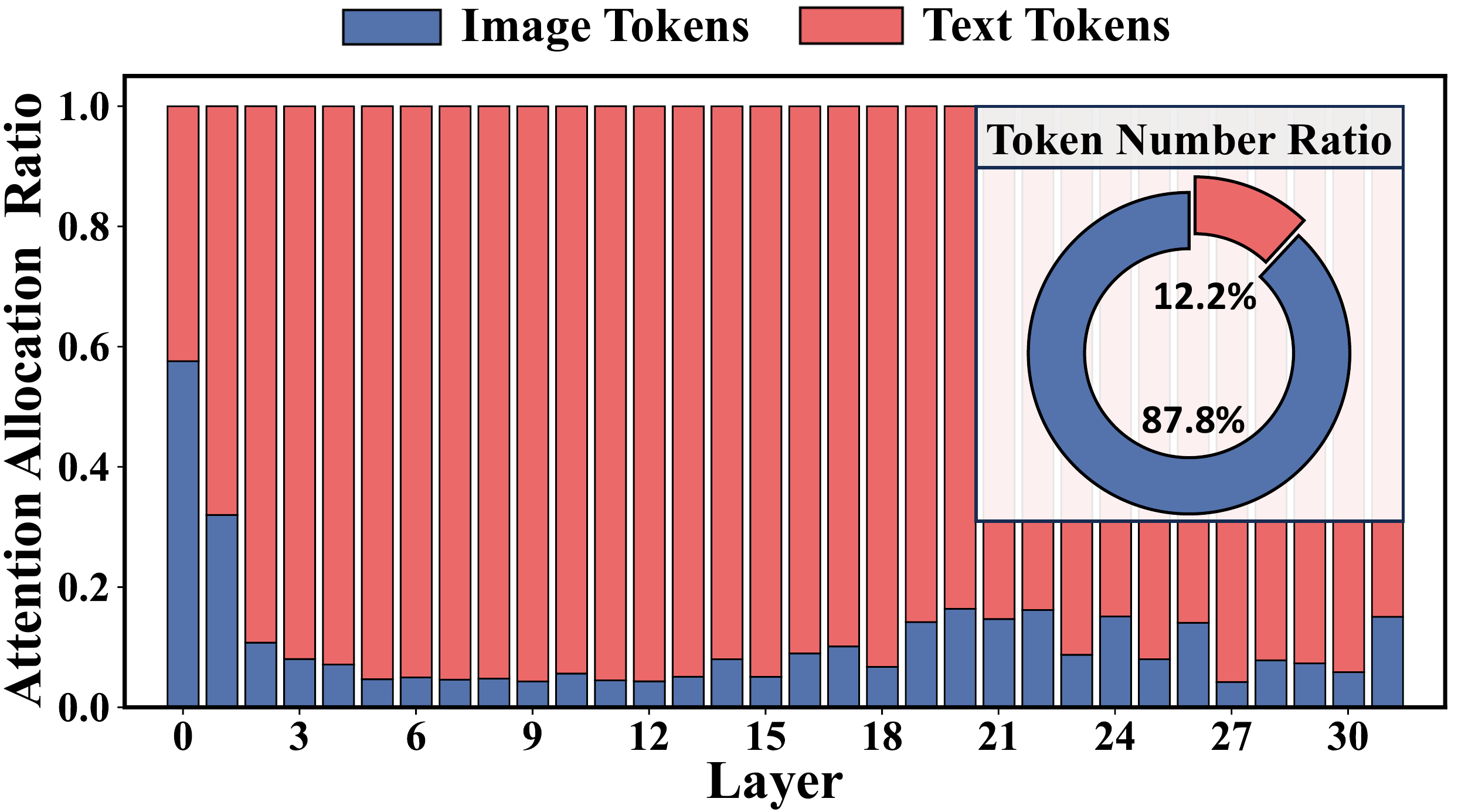} 

    \vspace{-2mm}
    \caption{Analysis of the input token number ratios ($r_{\text{num}}^{t}, r_{\text{num}}^{v}$) and attention allocation ratios ($r_{\text{att}}^{t}, r_{\text{att}}^{v}$). While image tokens comprise 87.8\% of the input sequence, their attention is disproportionately low.}
    \label{fig:attention_imbalance}
\end{figure}

 As illustrated in \autoref{fig:attention_imbalance}, the attention allocated to image tokens is markedly lower than that given to text tokens, despite image tokens constituting the vast majority ($r_{\text{num}}^{v} = 87.8\%$) of the input sequence. Specifically, with the exception of the two layers closest to the input, the attention allocated to image tokens in the subsequent layers ranges from 4.2\% to 16.4\%, significantly lower than that allocated to text tokens. This further validates that during inference, the LVLM's focus on visual input is insufficient, leading to an output distribution biased towards language priors.

Based on these observations, we assume that the primary cause of action-relation hallucinations in LVLMs is the insufficient attention allocated to visual information. 
Therefore, we posit that \textbf{enhancing the LVLM's attention towards action-relevant image tokens is essential to mitigate action-relation hallucinations.}
For instance, as illustrated in Figure~\ref{fig:action_hallucination}, it is essential to enhance the attention specifically on the hand regions, which are highly relevant to the ``pushing'' action verb.
To achieve this, we confront two key challenges:

\noindent \textit{Challenge 1: How to locate action-relevant image regions?} 
Prior research has revealed that attention heads in LLMs exhibit functional specialization, such as retrieval heads for extracting relevant information from long contexts~\cite{basile2025head,wu2024retrieval}. 
Inspired by these insights, we aim to identify the attention heads that are sensitive to action-relation changes. 
Subsequently, we locate the highly activated regions within these attention heads and select them as the action-relevant image regions. Please see Section~\ref{subsec:head_identification} for details.

\noindent \textit{Challenge 2: How to effectively steer the LVLM's focus toward action-relevant image regions?} 
We propose to amplify the attention weights corresponding to these action-relevant image regions, while reducing the interference of irrelevant background noise.  
This enables the model to focus more on action-relevant image information during inference, thereby mitigating action-relation hallucinations. Please see Section~\ref{subsec:visual_enhancement} for details.

\begin{figure*}[t]
    \centering
    \includegraphics[width=1.0\linewidth]{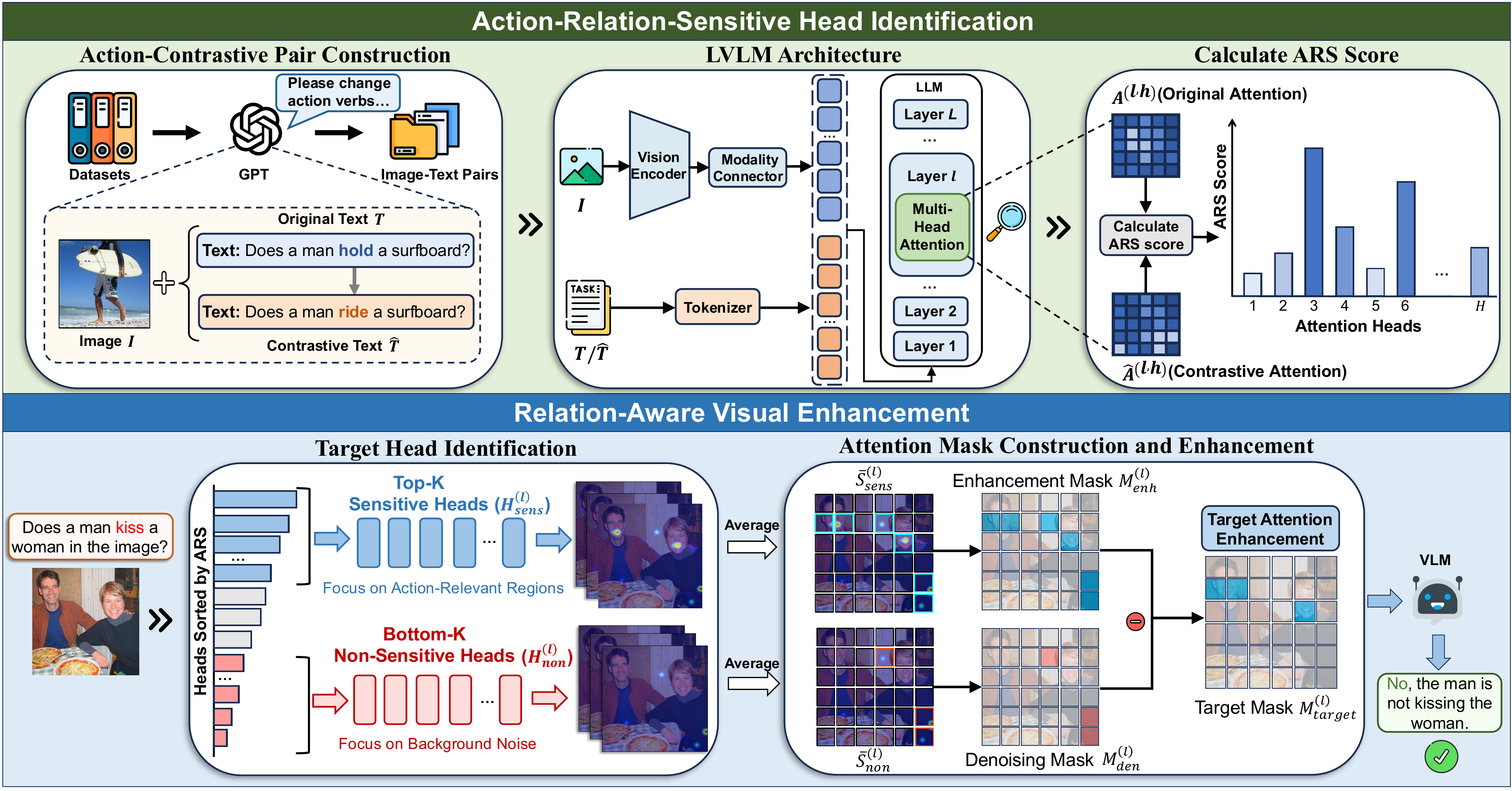}

    \caption{Overview of the proposed framework, consisting of two core components: (Top) Action-Relation-Sensitive Head Identification, which calculates ARS scores to identify  attention heads critical for action-relation reasoning; and (Bottom) Relation-Aware Visual Enhancement, which leverages these attention heads to enhance action-relevant image regions during inference.}
    \vspace{-2mm}
    \label{fig:framework}
\end{figure*}

\begin{figure}[t]
    \centering
    \includegraphics[width=1.0\linewidth]{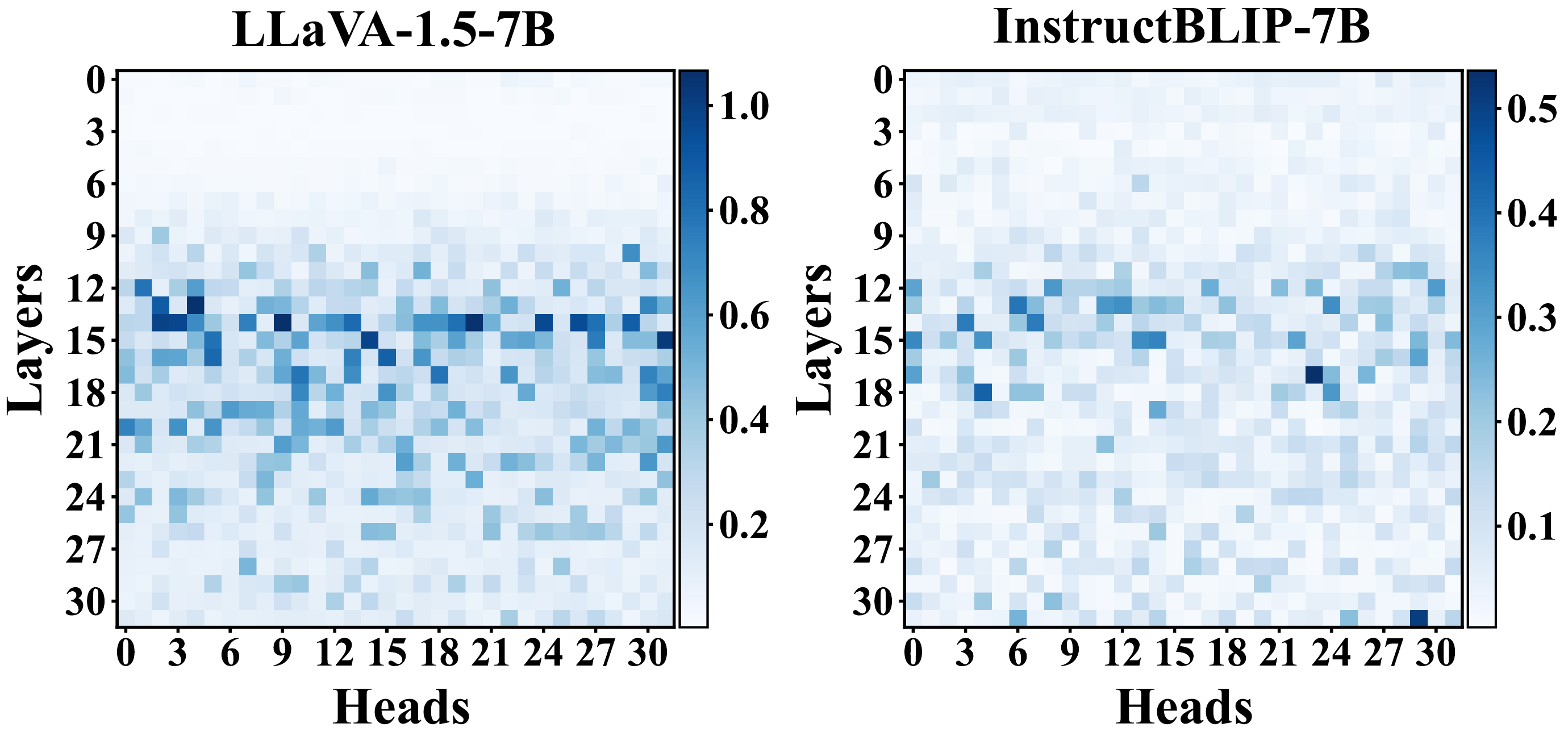}
    \caption{Distribution of ARS score values across layers and heads. Results indicate that the middle layers display high sensitivity to action-relation changes.}
    \label{fig:ARS_score}
\end{figure}

\begin{figure*}[t]
    \centering
    \includegraphics[width=1.0\linewidth]{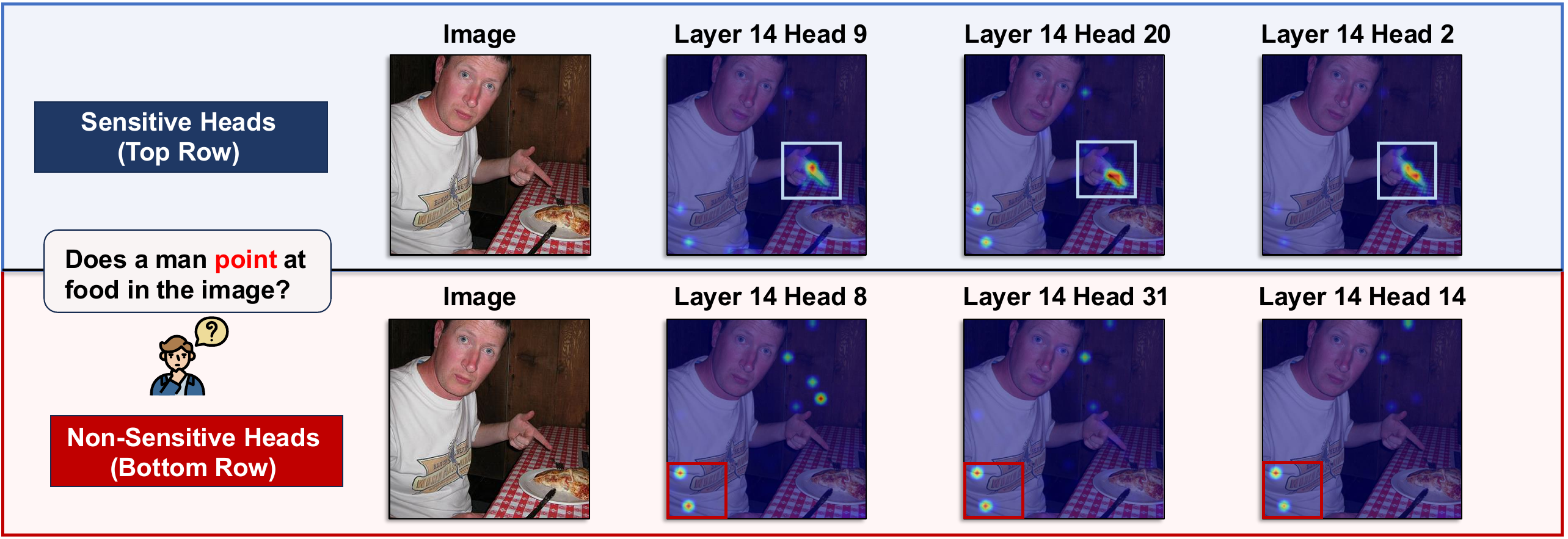}

    \vspace{-1.8mm}
    \caption{Visualization of attention maps at Layer 14 of LLaVA-1.5-7B. Sensitive heads (top) localize image regions relevant to the "point" action, while non-sensitive heads (bottom) focus on background noise, validating the effectiveness of ARS score.}
    \label{fig:Visualization_of_attention_maps}
    \vspace{-1.5mm}
\end{figure*}

\section{Method}
To address these two key challenges, we propose a training-free framework to mitigate action-relation hallucinations in LVLMs. 
As illustrated in Figure~\ref{fig:framework}, the proposed framework comprises two modules: the action-relation-sensitive head identification module (Section~\ref{subsec:head_identification}) and the relation-aware visual enhancement module (Section~\ref{subsec:visual_enhancement}).

\subsection{Action-Relation-Sensitive Head Identification}
\label{subsec:head_identification}

To identify the attention heads that are sensitive to action relations, we first construct action-contrastive pairs. 
Specifically, given an input pair consisting of an image $I$ and a text $T$, we generate a contrastive text $\hat{T}$ by substituting the action verb in $T$ with a semantically distinct alternative. 
For example, given the input text ``\emph{Does a man hold a surfboard},'' we change the verb ``\emph{hold}'' to ``\emph{ride},'' thereby creating a contrastive query, as shown in Figure~\ref{fig:framework}. 
We utilize GPT-5 to automatically generate these contrastive samples.
For further details regarding the generation process and more examples, please see the Appendix~\ref{sec:appendix_generation}. Through this process, we construct input pairs $(I, T)$ and $(I, \hat{T})$ that differ only in the verb. In this way, for the $h$-th head in the $l$-th layer, we obtain the visual attention weights $A^{(l,h)}$ and $\hat{A}^{(l,h)}$ corresponding to the inputs $(I, T)$ and $(I, \hat{T})$, respectively.

To evaluate the sensitivity of attention heads to action-relation changes, we propose the following metric to quantify the divergence in attention distributions, which is termed the Action-Relation Sensitivity (ARS) score. 
Formally, the ARS score for the $h$-th head in the $l$-th layer is defined as:
\begin{equation}
    \text{ARS}^{(l,h)} = \frac{\|A^{(l,h)} - \hat{A}^{(l,h)}\|_F}{\frac{1}{2}\left(\|A^{(l,h)}\|_F + \|\hat{A}^{(l,h)}\|_F\right)},
\end{equation}
where $\|\cdot\|_F$ denotes the Frobenius norm, and the denominator is introduced for normalization.
Figure \ref{fig:ARS_score} visualizes the distribution of ARS scores across all layers and attention heads. Results demonstrate a layer-wise pattern: attention heads within the middle layers exhibit notably higher ARS scores, whereas attention heads in shallow and deep layers display significantly lower scores. This observation suggests that the middle layers are more sensitive to action-relation changes. 

\noindent \textbf{Verifying the effectiveness of ARS scores.} 
We  visualize the attention maps of various heads within middle layers. If attention heads with high ARS scores indeed focus more on action-relevant image regions, then we can conclude that the ARS score effectively reflects the sensitivity to action-relation changes. Figure~\ref{fig:Visualization_of_attention_maps} displays the attention maps of the top-3 heads with the highest ARS scores and the bottom-3 heads with the lowest ARS scores in the middle layer. Results show that attention heads with the highest ARS scores focus on the image regions of the hand, which correspond to the action-relevant word ``\emph{point}.'' In contrast, the heads with the lowest ARS scores tend to focus on irrelevant regions. Such qualitative analysis verifies that the ARS score effectively reflects the sensitivity of attention heads to action-relation changes. Please see Section~\ref{subsec:ablation} for more verification of the effectiveness of ARS scores.

\begin{table*}[t]
\centering
\renewcommand{\arraystretch}{0.9} 
\setlength{\tabcolsep}{1.8pt}
\newcommand{\wbox}[1]{\makebox[3.5em]{#1}}

\resizebox{\linewidth}{!}{
\begin{tabular}{cl|cc|cc|cc|cc|cc}
\toprule
\multirow{2.5}{*}{\textbf{Benchmark}} & \multicolumn{1}{c|}{\multirow{2.5}{*}{\textbf{Method}}} & 
\multicolumn{2}{c|}{\textbf{LLaVA-1.5-7B}} & 
\multicolumn{2}{c|}{\textbf{LLaVA-NeXT-7B}} & 
\multicolumn{2}{c|}{\textbf{InstructBLIP-7B}} & 
\multicolumn{2}{c|}{\textbf{ShareGPT4V-7B}} & 
\multicolumn{2}{c}{\textbf{LLaVA-1.5-13B}} \\
\cmidrule(lr){3-4} \cmidrule(lr){5-6} \cmidrule(lr){7-8} \cmidrule(lr){9-10} \cmidrule(lr){11-12}
 & & \wbox{Acc} & \wbox{F1} & \wbox{Acc} & \wbox{F1} & \wbox{Acc} & \wbox{F1} & \wbox{Acc} & \wbox{F1} & \wbox{Acc} & \wbox{F1} \\
\midrule

\multirow{5}{*}{\shortstack[c]{\textbf{MMRel}\textsuperscript{\textdagger}\\\textbf{(Real)}}} 
  & Vanilla & 71.12 & 77.32 & 70.58 & 77.21 & 65.47 & 74.32 & 72.96 & 78.44 & \underline{74.51} & \underline{79.33} \\
  & + VCD        & 64.67 & 73.83 & 68.86 & 76.30 & \underline{68.97} & \underline{79.62} & \underline{73.15} & \underline{78.48} & 71.80 & 77.77 \\
  & + ICD        & \underline{73.30} & \underline{78.54} & \underline{70.79} & \underline{77.34} & 61.10 & 71.99 & 72.31 & 78.07 & 74.06 & 79.01 \\
  & + VAF        & 68.02 & 75.63 & 69.62 & 76.68 & 62.70 & 72.91 & 69.84 & 76.74 & 72.52 & 78.22 \\
  & + \textbf{Ours} & \textbf{78.08} & \textbf{81.10} & \textbf{77.39} & \textbf{81.23} & \textbf{78.56} & \textbf{79.63} & \textbf{76.44} & \textbf{80.47} & \textbf{78.36} & \textbf{81.73} \\
\cmidrule(lr){2-12} 

\multirow{5}{*}{\shortstack[c]{\textbf{MMRel}\\\textbf{(DALL-E)}}}
  & Vanilla & 69.06 & 73.28 & 67.54 & 74.67 & 69.65 & 75.19 & 69.06 & 74.86 & \underline{72.44} & \underline{77.55} \\
  & + VCD        & 62.97 & 72.45 & 62.46 & 72.86 & \underline{71.09} & \underline{75.43} & \underline{70.16} & \underline{75.05} & 67.54 & 75.26 \\
  & + ICD        & \underline{69.57} & 72.35 & \underline{67.62} & \underline{74.79} & 67.29 & 73.44 & 68.81 & 74.53 & 71.77 & 77.19 \\
  & + VAF        & 67.54 & \underline{73.84} & 66.78 & 74.36 & 68.47 & 75.31 & 67.79 & 74.31 & 70.84 & 77.02 \\
  & + \textbf{Ours} & \textbf{71.60} & \textbf{75.07} & \textbf{70.58} & \textbf{75.00} & \textbf{72.78} & \textbf{75.64} & \textbf{72.53} & \textbf{75.62} & \textbf{75.23} & \textbf{78.31} \\
\midrule

\multirow{5}{*}{\shortstack[c]{\textbf{R-Bench}\textsuperscript{\textdagger}\\\textbf{(Image)}}}
  & Vanilla & 78.05 & 85.34 & \underline{83.05} & \underline{87.83} & \underline{81.74} & \underline{87.31} & 80.47 & 86.73 & 82.76 & 87.38 \\
  & + VCD        & \underline{78.62} & \underline{85.54} & 82.50 & 87.49 & 81.35 & 86.87 & \underline{81.40} & \underline{86.98} & 81.36 & 86.23 \\
  & + ICD        & 72.74 & 82.26 & 82.98 & 87.76 & 81.66 & 87.20 & 74.80 & 83.22 & 82.55 & 87.20 \\
  & + VAF        & 77.69 & 85.23 & 82.89 & 87.76 & 81.38 & 87.24 & 79.81 & 86.38 & \underline{82.89} & \underline{87.53} \\
  & + \textbf{Ours} & \textbf{79.10} & \textbf{85.64} & \textbf{83.50} & \textbf{88.13} & \textbf{82.48} & \textbf{87.59} & \textbf{81.84} & \textbf{87.45} & \textbf{83.26} & \textbf{87.76} \\
\cmidrule(lr){2-12}

\multirow{5}{*}{\shortstack[c]{\textbf{R-Bench}\\\textbf{(Instance)}}}
  & Vanilla & 62.94 & 73.45 & 64.59 & 83.15 & 66.70 & 77.57 & 66.37 & 76.54 & 69.12 & 79.37 \\
  & + VCD        & \underline{68.65} & \underline{79.76} & 72.87 & 83.03 & 67.30 & 78.15 & \underline{69.77} & \underline{79.77} & 70.05 & 80.43 \\
  & + ICD        & 59.22 & 69.83 & 68.50 & 78.80 & \underline{68.53} & \underline{79.33} & 66.25 & 76.40 & 69.01 & 79.30 \\
  & + VAF        & 67.69 & 77.61 & \underline{73.22} & \underline{83.35} & 68.05 & 79.00 & 68.95 & 79.10 & \underline{70.45} & \underline{80.67} \\
  & + \textbf{Ours} & \textbf{74.46} & \textbf{84.97} & \textbf{74.46} & \textbf{84.56} & \textbf{70.64} & \textbf{81.20} & \textbf{75.86} & \textbf{86.00} & \textbf{75.73} & \textbf{85.60} \\
\midrule

\multirow{5}{*}{\shortstack[c]{\textbf{AMBER}\\\textbf{(Relation)}}}
  & Vanilla & 58.41 & 45.43 & 62.62 & 53.86 & 68.87 & 75.26 & 67.13 & 61.61 & 73.44 & 71.92 \\
  & + VCD        & 60.46 & 50.75 & 55.83 & 39.90 & \underline{69.41} & 76.24 & 69.83 & 66.67 & 73.32 & 72.28 \\
  & + ICD        & 59.38 & 47.43 & 61.06 & 50.91 & 68.39 & \underline{77.09} & 68.09 & 63.15 & 73.20 & 71.63 \\
  & + VAF        & \underline{62.86} & \underline{54.63} & \underline{66.47} & \underline{60.65} & 68.81 & 76.86 & \underline{70.79} & \underline{67.69} & \underline{73.56} & \underline{72.60} \\
  & + \textbf{Ours} & \textbf{71.81} & \textbf{74.85} & \textbf{75.24} & \textbf{75.21} & \textbf{73.32} & \textbf{78.80} & \textbf{76.86} & \textbf{76.60} & \textbf{81.25} & \textbf{83.32} \\
\bottomrule
\end{tabular}%
}
\begin{minipage}{\linewidth}
\footnotesize
\textsuperscript{\textdagger} We focus our evaluation on the action-relation subsets for these benchmarks. For spatial-relation, please refer to Table~\ref{tab:spatial_results}.
\end{minipage}

\vspace{-4pt}
\caption{Performance comparison of vanilla LVLMs and different training-free mitigation methods on action-relation hallucinations across MMRel, R-Bench, and AMBER benchmarks. Please see Appendix~\ref{sec:appendix_case_studies} for case studies.}
\label{tab:main_results}
\end{table*}

\subsection{Relation-Aware Visual Enhancement}
\label{subsec:visual_enhancement}
Building on the previous analysis, we propose the RVE method to amplify the LVLM's attention towards action-relevant image regions by modifying the pre-softmax attention scores $S^{(l,h)}$. As illustrated in Figure~\ref{fig:Visualization_of_attention_maps} (top), the action-relevant regions are primarily identified by the action-relation-sensitive heads. We also observe that action-relation-sensitive heads capture not only action-relevant regions (e.g., \textit{fingers}) but also action-irrelevant regions (e.g., \textit{clothing}). Notably, these irrelevant regions are also captured by non-sensitive heads, as shown in Figure~\ref{fig:Visualization_of_attention_maps} (bottom). To avoid enhancing these irrelevant regions, we construct two attention masks to determine which image tokens require enhancement and which should be excluded.

\noindent \textbf{Attention mask construction.} 
Let $H^{(l)}_{\text{sens}}(K)$ denote the set of the top-$K$ heads with the highest ARS scores, termed \textit{sensitive heads}, and $H^{(l)}_{\text{non}}(K)$ denote the set of the bottom-$K$ heads with the lowest ARS scores, termed \textit{non-sensitive heads}.
Based on these sets, we compute the averaged attention scores for the sensitive heads as $\bar{S}_{\text{sens}}^{(l)} = \mathbb{E}_{h \in H_{\text{sens}}^{(l)}} [ S^{(l,h)} ] \in \mathbb{R}^{1 \times N_I}$, and for the non-sensitive heads as $\bar{S}_{\text{non}}^{(l)} = \mathbb{E}_{h \in H_{\text{non}}^{(l)}} [ S^{(l,h)} ] \in \mathbb{R}^{1 \times N_I}$. Then, we construct an enhancement mask $M_{\text{enh}}^{(l)} \in \{0, 1\}^{N_I}$ and a denoising mask $M_{\text{den}}^{(l)} \in \{0, 1\}^{N_I}$ as follows:
\begin{equation}
    \begin{aligned}
        M_{\text{enh}}^{(l)}[i] &= \begin{cases} 1, & \text{if } \bar{S}_{\text{sens}}^{(l)}[i] \ge \tau_{\text{sens}} \\ 0, & \text{otherwise} \end{cases}, \\
        M_{\text{den}}^{(l)}[i] &= \begin{cases} 1, & \text{if } \bar{S}_{\text{non}}^{(l)}[i] \ge \tau_{\text{non}} \\ 0, & \text{otherwise} \end{cases},
    \end{aligned}
\end{equation}
where $\tau_{\text{sens}}$ and $\tau_{\text{non}}$ are the values of the $m$-th largest elements in $\bar{S}_{\text{sens}}^{(l)}$ and $\bar{S}_{\text{non}}^{(l)}$, respectively. Here, $m = \lfloor \alpha \cdot N_I \rfloor$ is the number of selected tokens determined by the ratio $\alpha$. In this way, the enhancement mask and the denoising mask are constructed to select the top-$m$ image tokens in $\bar{S}_{\text{sens}}^{(l)}$ and $\bar{S}_{\text{non}}^{(l)}$, respectively.

To precisely locate action-relevant visual cues while avoiding the interference of background noise, we define the target mask $M_{\text{target}}^{(l)}$ as follows:
\begin{equation}
    M_{\text{target}}^{(l)} = M_{\text{enh}}^{(l)} \odot (1 - M_{\text{den}}^{(l)}),
\end{equation}
where $\odot$ denotes the element-wise multiplication. 

As shown in Figure~\ref{fig:framework} (bottom), the enhancement mask $M_{\text{enh}}^{(l)}$ identifies both \textit{the mouth regions} and \textit{three background regions} as candidates for attention enhancement. Meanwhile, the denoising mask $M_{\text{den}}^{(l)}$ identifies these \textit{three background regions} as noise that should not be enhanced. In this way, the target mask $M_{\text{target}}^{(l)}$ exclusively enhances the regions truly relevant to the action relation (i.e., \textit{the mouth regions}) without amplifying action-irrelevant image regions.

\noindent \textbf{Relation-aware visual enhancement.}
Finally, for each attention head $h$ in layer $l$, we utilize the target mask $M_{\text{target}}^{(l)}$ to enhance the attention scores towards action-relevant image regions:
\begin{equation}
    \tilde{S}^{(l,h)} = S^{(l,h)} + \beta \cdot (\lvert S^{(l,h)} \rvert \odot M_{\text{target}}^{(l)}),
    \label{eq:rve}
\end{equation}
where $\beta > 0$ is the enhancement coefficient. 

\section{Experiments}
\noindent \textbf{LVLMs and Benchmarks.} We conduct experiments on \textbf{five LVLMs}: LLaVA-1.5-7B, LLaVA-1.5-13B~\cite{liu2024improved}, LLaVA-NeXT-7B~\cite{liu2024llavanext}, ShareGPT4V-7B~\cite{chen2024sharegpt4v}, and InstructBLIP-7B~\cite{dai2023instructblip}. To comprehensively evaluate the effectiveness and generalizability of our method, we conduct experiments under \textbf{four evaluation scenarios}: discriminative action-relation hallucination, generative action-relation hallucination, spatial-relation hallucination, and object hallucination.
We compare our RVE method with \textbf{four benchmarks}: MMRel~\cite{nie2024mmrel}, R-Bench~\cite{wu2024evaluating}, and AMBER~\cite{wang2023amber} for relation hallucination evaluation, and POPE~\cite{li2023evaluating} for object hallucination evaluation.
For each scenario, the specific benchmarks and their corresponding subsets are presented in Tables~\ref{tab:main_results}, \ref{tab:open_ended_results}, \ref{tab:spatial_results}, and \ref{tab:pope_results}. For detailed descriptions of these subsets, please refer to Appendix~\ref{sec:appendix_benchmarks}.

\noindent \textbf{Baselines and Experimental Settings.} We compare our method against vanilla LVLMs and three training-free mitigation methods (VCD~\cite{leng2024mitigating}, ICD~\cite{wang2024mitigating}, and VAF~\cite{yin2025clearsight}) applied to each vanilla model.
Note that these three mitigation methods are originally designed for object hallucinations.
To the best of our knowledge, there are currently no methods specifically designed for mitigating action-relation hallucinations. Regarding hyperparameters, we uniformly set $K=5$ and $\beta=1.0$ across all models. We set $\alpha$ to $0.05$ for the LLaVA series and ShareGPT4V, and $0.5$ for InstructBLIP. Please see Section~\ref{subsec:ablation} for detailed analysis.

\begin{table}[t]
    \centering
    \renewcommand{\arraystretch}{1.0} 
    \setlength{\tabcolsep}{3.5pt}       
    
    \resizebox{\linewidth}{!}{
    \begin{tabular}{l|cc|cc}
        \toprule
        \multirow{2}{*}{\textbf{Model}} & \multicolumn{2}{c|}{\textbf{MMRel (Real)}} & \multicolumn{2}{c}{\textbf{MMRel (DALL-E)}} \\
        \cmidrule(lr){2-3} \cmidrule(lr){4-5}
         & \textbf{Vanilla} & \textbf{Ours} & \textbf{Vanilla} & \textbf{Ours} \\
        \midrule
        LLaVA-1.5-7B    & 7.84 & \textbf{7.91} & 6.76 & \textbf{6.89} \\
        LLaVA-NeXT-7B   & 8.15 & \textbf{8.21} & 7.15 & \textbf{7.32} \\
        InstructBLIP-7B & 6.19 & \textbf{6.28} & 7.41 & \textbf{7.47} \\
        ShareGPT4V-7B   & 8.18 & \textbf{8.26} & 6.87 & \textbf{7.06} \\
        \bottomrule
    \end{tabular}
    }
    \caption{LLM-assisted open-ended evaluation on MMRel (Real) and MMRel (DALL-E).}
    \label{tab:open_ended_results}
    \vspace{-2.5mm}
\end{table}

\subsection{Main Results}

\noindent \textbf{Performance on action-relation hallucinations.} Results in Table~\ref{tab:main_results} demonstrate that our method achieves consistent performance gains across all evaluated LVLMs and benchmarks, with \textit{relative accuracy improvements} over the vanilla baseline ranging from 0.54\% to 22.94\%. Notably, our RVE method generalizes well across real and synthetic domains on MMRel, while maintaining effectiveness across different granularities on R-Bench. Furthermore, the consistent gains observed on both 7B and 13B models demonstrate the scalability and generalization of our RVE method.

\noindent \textbf{Performance on open-ended generation tasks.} 
In addition to discriminative metrics (accuracy and F1 scores), we evaluate the generative quality of the models using GPT-5-mini scoring (scale 0--10); please see Appendix~\ref{sec:appendix_eval_details} for details. As Table~\ref{tab:open_ended_results} shows, our method achieves higher quality scores across all architectures compared to the vanilla baseline. This demonstrates that our method effectively mitigates hallucinations without degrading the LVLM's generation capability.

\noindent \textbf{Verifying generalization to spatial-relation hallucinations.} As shown in Table~\ref{tab:spatial_results}, our method consistently outperforms the vanilla baseline across all LVLMs on spatial-relation hallucinations. 
Notably, our method achieves consistent performance gains with \textit{relative improvements} over the vanilla baseline ranging from 0.80\% to 23.57\%, validating the generalization capability of our method to spatial-relation hallucinations.

\begin{table}[t]
\centering
\renewcommand{\arraystretch}{1.0}
\setlength{\tabcolsep}{3pt}

\resizebox{1.0\linewidth}{!}{
    \small 
    \begin{tabular}{ll|cc|cc}
    \toprule
    \multirow{2}{*}{\textbf{Model}} & \multirow{2}{*}{\textbf{Method}} & \multicolumn{2}{c|}{\textbf{MMRel}} & \multicolumn{2}{c}{\textbf{R-Bench}} \\
    \cmidrule(lr){3-4} \cmidrule(lr){5-6}
     & & \textbf{Real} & \textbf{DALL-E} & \textbf{Image} & \textbf{Instance} \\
    \midrule
    \multirow{2}{*}{LLaVA-1.5-7B} & Vanilla & 50.25 & 50.41 & 78.21 & 60.85 \\
     & + \textbf{Ours} & \textbf{54.61} & \textbf{54.05} & \textbf{79.08} & \textbf{75.19} \\
    \midrule
    \multirow{2}{*}{LLaVA-NeXT-7B} & Vanilla & 52.49 & 50.58 & 79.35 & 63.51 \\
     & + \textbf{Ours} & \textbf{54.22} & \textbf{54.30} & \textbf{81.13} & \textbf{76.29} \\
    \midrule
    \multirow{2}{*}{InstructBLIP-7B} & Vanilla & 49.90 & 51.82 & 77.62 & 69.12 \\
     & + \textbf{Ours} & \textbf{54.02} & \textbf{54.63} & \textbf{79.21} & \textbf{77.39} \\
    \midrule
    \multirow{2}{*}{ShareGPT4V-7B} & Vanilla & 53.01 & 53.31 & 79.67 & 64.61 \\
     & + \textbf{Ours} & \textbf{57.35} & \textbf{55.37} & \textbf{80.31} & \textbf{77.30} \\
    \bottomrule
    \end{tabular}
}

\vspace{-1mm}
\caption{Verifying the generalization of our RVE method to spatial-relation hallucinations.}
\label{tab:spatial_results}
\end{table}

\begin{table}[t]
\centering
\renewcommand{\arraystretch}{0.95} 
\setlength{\tabcolsep}{3pt}       

\resizebox{0.95\linewidth}{!}{
    \small 
    \begin{tabular}{l|cc|cc|cc}
    \toprule
    \multirow{2}{*}{\textbf{Method}} & \multicolumn{2}{c|}{\textbf{Random}} & \multicolumn{2}{c|}{\textbf{Popular}} & \multicolumn{2}{c}{\textbf{Adversarial}} \\
    \cmidrule(lr){2-3} \cmidrule(lr){4-5} \cmidrule(lr){6-7}
     & \textbf{Acc} & \textbf{F1} & \textbf{Acc} & \textbf{F1} & \textbf{Acc} & \textbf{F1} \\
    \midrule
    Vanilla & 87.07 & 85.50 & 85.83 & 84.33 & 82.97 & 81.26 \\
    VCD & \underline{88.37} & \textbf{87.91} & 86.20 & 85.98 & 83.33 & 82.02 \\
    VAF & 88.30 & 87.29 & \underline{86.77} & \underline{86.11} & \textbf{83.72} & \textbf{83.05} \\
    \textbf{Ours} & \textbf{88.63} & \underline{87.51} & \textbf{87.17} & \textbf{86.59} & \underline{83.65} & \underline{82.45} \\
    \bottomrule
    \end{tabular}
}
\vspace{-1mm}
\caption{Verifying the generalization of our RVE method to object hallucination. We report results on the three negative sampling settings of the POPE benchmark evaluated on MS-COCO, using LLaVA-1.5-7B.}
\label{tab:pope_results}
\vspace{-2.0mm}
\end{table}

\noindent \textbf{Verifying generalization to object hallucinations.} While our RVE method is designed to mitigate action-relation hallucinations, we further explore its generalizability to object hallucinations. As Table~\ref{tab:pope_results} shows, our method achieves competitive or superior performance on the POPE benchmark against the vanilla baseline and other object-hallucination mitigation methods. These findings effectively validate the generalization capability of our method in mitigating object hallucinations.

\subsection{Ablation and Further Analysis}
\label{subsec:ablation}

\begin{table}[t]
    \centering
    \renewcommand{\arraystretch}{0.9}
    \setlength{\tabcolsep}{3.5pt}
    
    \resizebox{\linewidth}{!}{
    \begin{tabular}{ccc|cc|cc}
        \toprule
        \multicolumn{3}{c|}{\textbf{Strategy}} & \multicolumn{2}{c|}{\textbf{LLaVA-1.5}} & \multicolumn{2}{c}{\textbf{InstructBLIP}} \\
        
        \textbf{Global} & \textbf{Sens.} & \textbf{Denoise} & \textbf{Acc} & \textbf{F1} & \textbf{Acc} & \textbf{F1} \\
        \midrule
        
         & & & 
        71.12 & 77.32 & 65.47 & 74.32 \\
        
        \midrule
        
         & \checkmark & & 
        73.78 & 78.78 & 67.29 & 74.88 \\
        
        \checkmark & & & 
        74.88 & 79.37 & 69.47 & 74.65 \\
        
        \midrule
        
         & \checkmark & \checkmark & 
        74.06 & 78.98 & 70.12 & 76.79 \\

        \checkmark & & \checkmark & 
        \textbf{78.08} & \textbf{81.10} & \textbf{78.56} & \textbf{79.63} \\
        
        \bottomrule
    \end{tabular}
    }
    
    \caption{Ablation of different enhancement scopes (global and sensitive-only) and the denoising mask on the real image subset of MMRel. See Appendix~\ref{sec:appendix_ablation} for ablations on more models.}
    \label{tab:ablation_results}
    \vspace{-2mm}
\end{table}

\begin{figure}[t]
    \centering
    \includegraphics[width=1.0\linewidth]{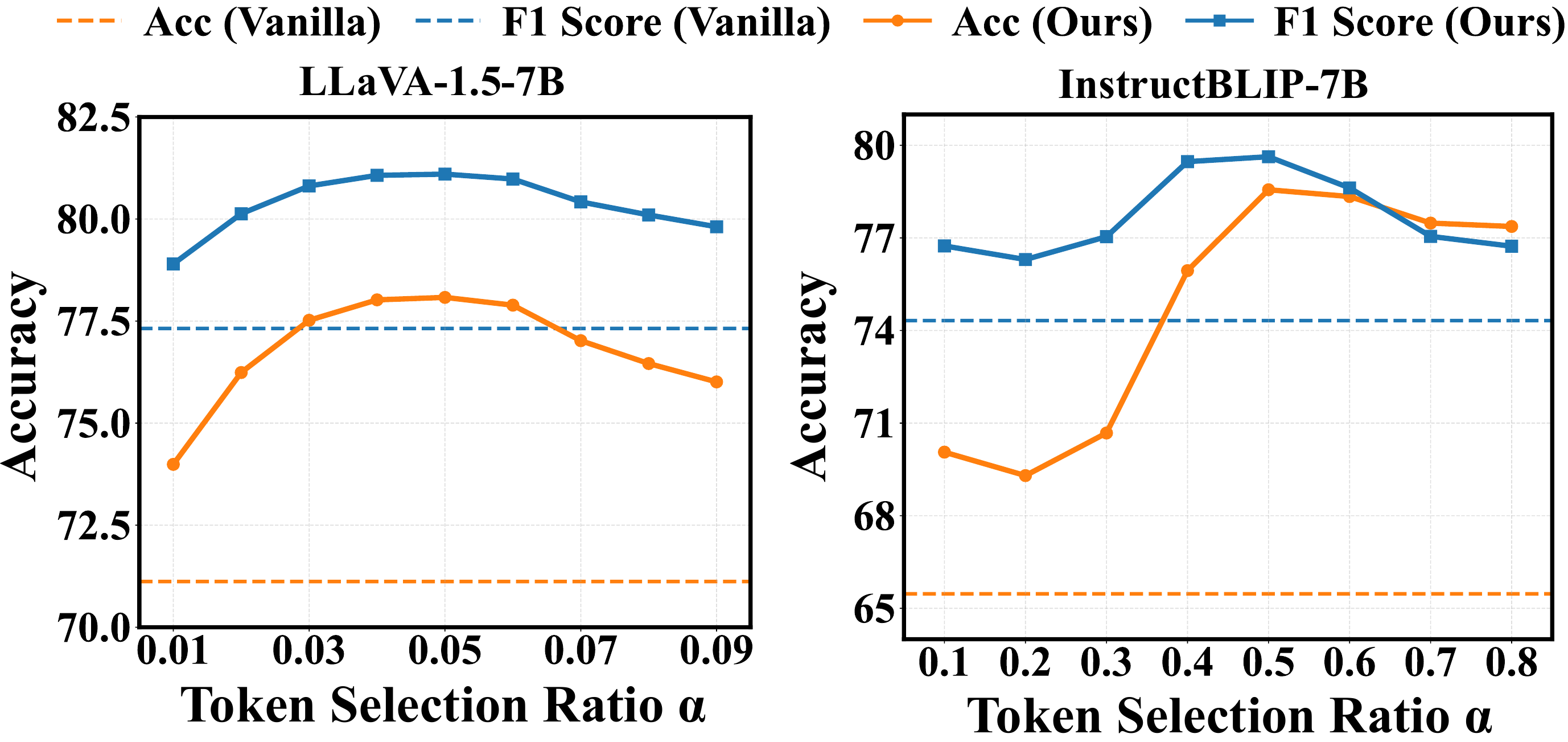}

    \vspace{-1.5mm}
    \caption{Impact of the selection ratio $\alpha$ on LLaVA-1.5-7B and InstructBLIP-7B evaluated on the real image subset of MMRel.}
    \label{fig:ablation}
    \vspace{-1.0mm}
\end{figure}

\noindent \textbf{Impact of global enhancement.} 
In Equation~\eqref{eq:rve}, we apply our RVE method to all attention heads in the target layer. In this experiment, we explore whether enhancing all heads (termed \textit{Global}) is more effective than enhancing only the sensitive heads (termed \textit{Sensitive-only}), where we select the top 50\% of heads ranked by the ARS score. As Table~\ref{tab:ablation_results} shows, under identical conditions, the \textit{Global} enhancement consistently outperforms \textit{Sensitive-only}. This indicates that aligning non-sensitive heads with sensitive ones further strengthens the LVLM’s focus on action-relevant regions.

\noindent \textbf{Impact of denoising mask.} 
As Table~\ref{tab:ablation_results} shows, our RVE method with the denoising mask consistently exhibits higher accuracy than the method without it. This proves the effectiveness of the denoising mask in preventing attention from being diverted to action-irrelevant regions.

\begin{figure}[t]
    \centering
    \includegraphics[width=1.0\linewidth]{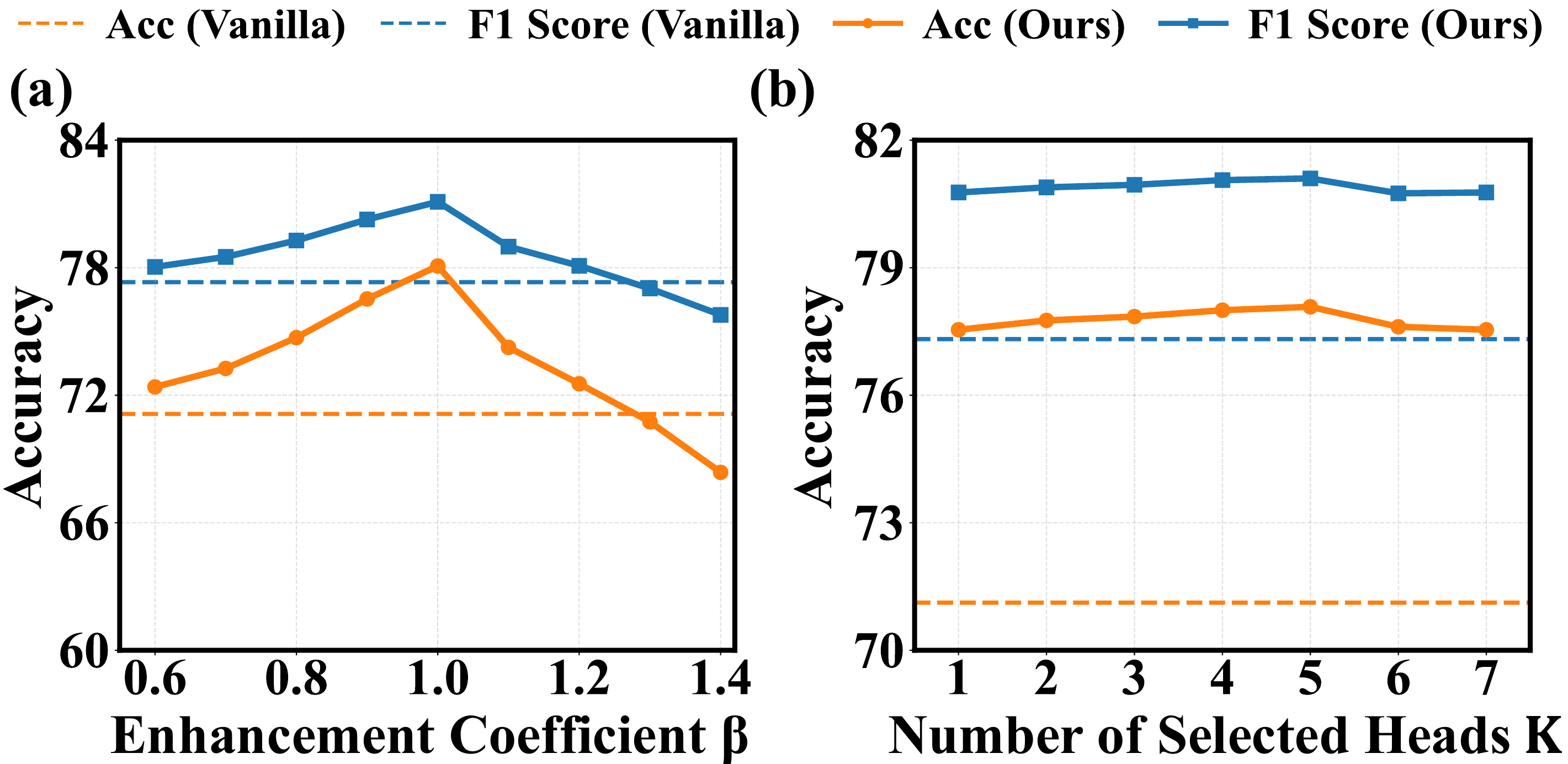}

    \vspace{-1mm}
    \caption{Results on LLaVA-1.5-7B. (a) Impact of the enhancement coefficient $\beta$ on performance. (b) Impact of the number of selected
    heads $K$ on performance.}
    \label{fig:beta_k}
    \vspace{-1.5mm}
\end{figure}

\noindent \textbf{Impact of hyperparameters.} 
Figure~\ref{fig:ablation} illustrates sensitivity to the selection ratio $\alpha$ across different architectures. 
For LLaVA-1.5-7B, the performance peaks at $\alpha=0.05$, indicating that a small fraction of critical tokens suffices to capture action semantics given its large number of image tokens. 
In contrast, InstructBLIP-7B requires a significantly higher selection ratio, achieving optimal performance at $\alpha=0.5$. We attribute this discrepancy to the Q-Former compressing visual features into only 32 image tokens, thus requiring a larger retention proportion to preserve sufficient action information.
Regarding the enhancement coefficient $\beta$, as shown in Figure~\ref{fig:beta_k} (a), performance steadily improves to a peak at $\beta=1.0$. A gradual decline is observed thereafter, suggesting that excessive enhancement distorts the original feature distribution and diminishes the LVLM's attention to necessary contextual information.
As Figure~\ref{fig:beta_k} (b) shows, performance reaches a peak when the number of selected heads $K=5$. Notably, the performance remains stable across different values of $K$ and consistently outperforms the vanilla baseline, validating the effectiveness of our ARS score in identifying action-relevant heads.

\begin{figure}[t]
    \centering
    \includegraphics[width=1.0\linewidth]{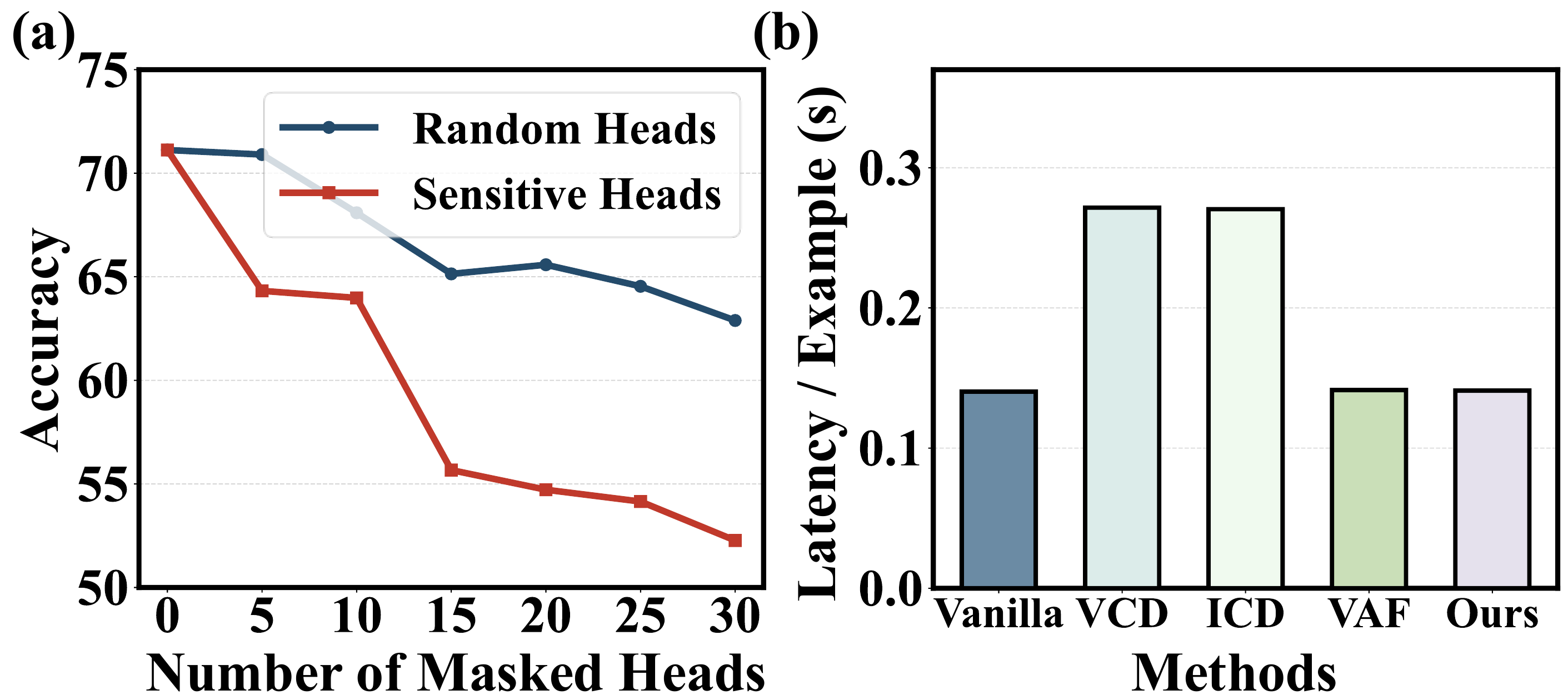}

    \vspace{-1mm}
    \caption{(a) Comparison of masking sensitive heads and random heads. (b) Inference latency per example across different methods.}
    \label{fig:mask_latency}
    \vspace{-1.5mm}
\end{figure}

\noindent \textbf{Verifying the effectiveness of selected sensitive heads.} 
As Figure~\ref{fig:mask_latency} (a) shows, masking randomly selected heads results in a slow performance decline, whereas masking the sensitive heads identified by the ARS score leads to a sharp accuracy drop. This significant gap confirms that the ARS score effectively locates the specific heads critical for action-relation understanding.

\noindent \textbf{Inference speed.} 
We further compare the inference latency of our method with baselines in Figure~\ref{fig:mask_latency} (b). Methods like VCD and ICD double latency due to additional computational requirements. In contrast, our method maintains a speed comparable to the vanilla baseline, demonstrating that our method achieves effective hallucination mitigation with negligible extra inference cost.

\section{Related Works}
\textbf{Large Vision-Language Models.} LVLMs have achieved remarkable success in multimodal tasks~\cite{caffagni2024revolution,li2025survey}. These models typically comprise a vision encoder, a modality connector, and a pretrained LLM. Representative models employ diverse alignment strategies, including MLP-based projectors in LLaVA-1.5~\cite{liu2024improved} and Q-Formers in InstructBLIP~\cite{dai2023instructblip}. Despite their impressive capabilities, these LVLMs still suffer from severe hallucinations, limiting their reliability in real-world applications.

\noindent\textbf{Hallucination in LVLMs.} Hallucination in LVLMs refers to inconsistencies between visual inputs and generated responses~\cite{he2025cracking}. Several approaches aim to mitigate hallucinations via reinforcement learning~\cite{zhao2023beyond,yu2024rlhf}, data refinement~\cite{Yu_2024_CVPR}, or post-hoc revisers~\cite{zhou2023analyzing}. However, these methods incur high training costs. In contrast, training-free methods like OPERA~\cite{huang2024opera} and Contrastive Decoding (e.g., VCD~\cite{leng2024mitigating}, ICD~\cite{wang2024mitigating}) offer lightweight solutions by adjusting attention or decoding distributions during inference. Nevertheless, existing methods mostly focus on mitigating object hallucinations, overlooking the more intricate issue of action-relation hallucinations~\cite{zheng2025reefknot,wu2024evaluating}. Consequently, we propose a training-free framework to mitigate action-relation hallucinations.

\section{Conclusion}
In this paper, we propose a training-free framework to mitigate action-relation hallucinations in LVLMs. Specifically, we define the ARS score to identify action-relation-sensitive attention heads and propose the RVE method to enhance attention toward action-relevant image regions. Extensive experiments across multiple benchmarks and LVLMs demonstrate that our method not only outperforms baselines in mitigating action-relation hallucinations but also effectively generalizes to spatial-relation and object hallucinations, with negligible additional inference cost.

\section*{Limitations}
While our proposed framework effectively mitigates action-relation hallucinations with negligible additional inference cost, it exhibits certain limitations. First, since our method involves directly adjusting the LVLM's attention, it requires access to the LVLM's internal layers and representations. This restricts its applicability to closed-source models  where only API-level access is available. Second, the optimal selection of layers for enhancement may vary across different model architectures and tasks to achieve optimal performance. Developing adaptive mechanisms that automatically determine the target layers based on dynamic metrics, such as generation confidence or attention entropy, remains a promising direction for future research.

\section*{Broader Impact and Ethics Statement}
Our research focuses on mitigating action-relation hallucinations to enhance the reliability and truthfulness of LVLMs. We evaluate our method using publicly available datasets and LVLMs across multiple hallucination-related benchmarks. While our method demonstrates promising results, its effectiveness is constrained by the inherent capabilities of the base model, and improper usage may adversely affect the model's performance. To the best of our knowledge, our proposed method does not introduce additional ethical concerns regarding data privacy or social bias.

\section*{Acknowledgements}
This work was partially supported by the National Natural Science Foundation of China (Nos. 62576249, 62376199, 62576247) and the Shanghai Municipal Education Commission (No. 24CGA20).

\bibliography{custom}

\appendix
\clearpage
\newpage

\section{Comparison of Difficulty between Object and Action-Relation Hallucinations}
\label{sec:appendix_gap_analysis}
\begin{figure}[h]
    \centering
    \includegraphics[width=0.9\linewidth]{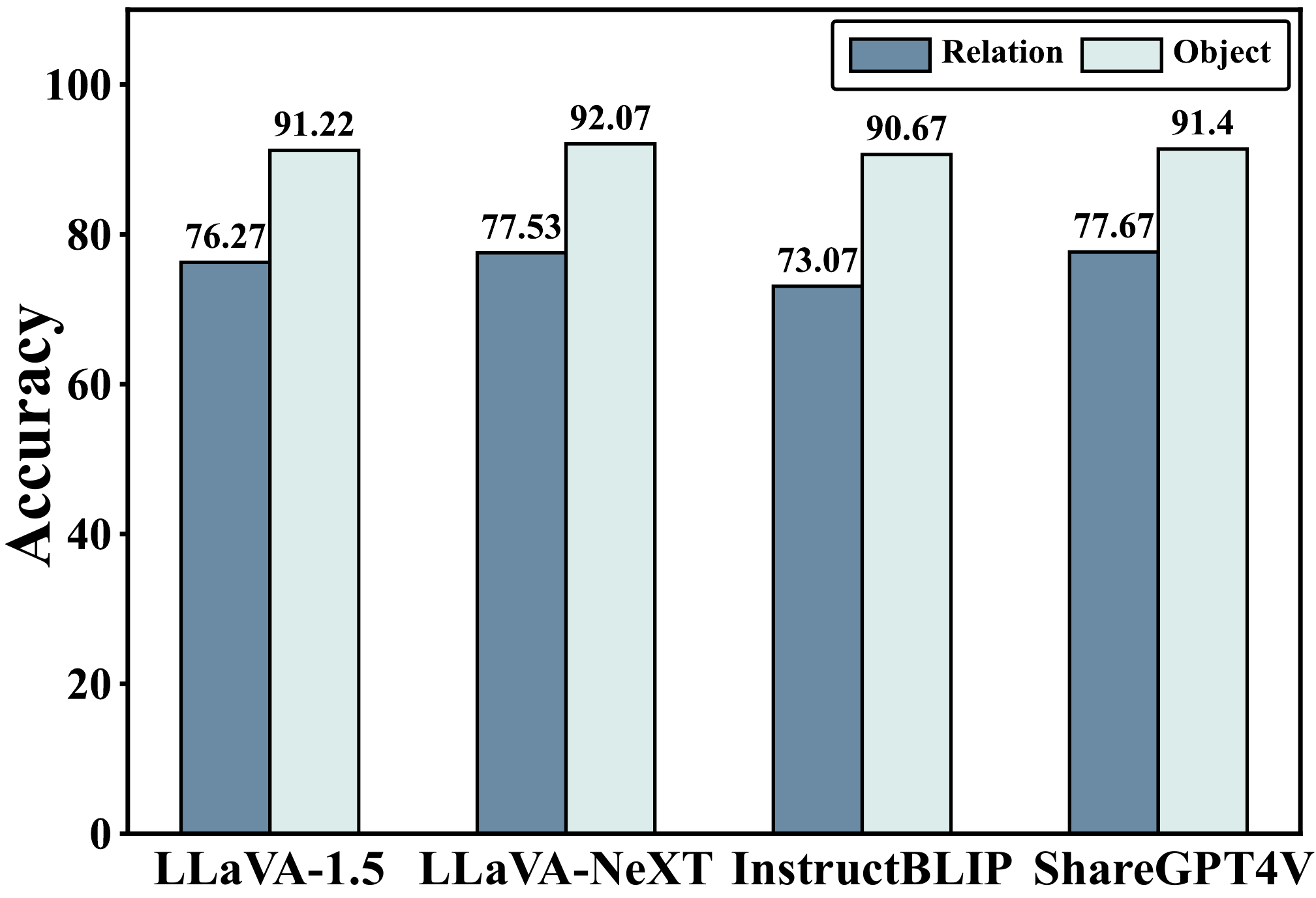}
    \caption{Performance comparison between object and action-relation hallucinations across four LVLMs.}
    \label{fig:motivation_gap}
\end{figure}

To quantitatively assess the performance disparity between object and action-relation understanding, we conduct a comparative analysis based on the R-Bench~\cite{wu2024evaluating} dataset. We specifically utilize 1,500 images, each accompanied by paired queries designed to evaluate object hallucination and action-relation hallucination, respectively. This paired experimental design ensures that both tasks are evaluated under identical visual contexts, effectively eliminating potential biases arising from image complexity or content distribution. As illustrated in \autoref{fig:motivation_gap}, the results reveal a significant performance gap: while the accuracy on object queries consistently exceeds 90\% across all models, the performance on action-relation hallucination lags behind by approximately 15\%. This observation confirms that action-relation hallucination represents a more severe challenge that necessitates dedicated mitigation strategies.

\section{Generation of Action-Contrastive Image-Text Pairs}
\label{sec:appendix_generation}

In this section, we describe the pipeline for generating action-contrastive image-text pairs. 

\subsection{Generation Pipeline}
To construct valid contrastive pairs $(I, \hat{T})$ from the original pairs $(I, T)$, we employ a systematic three-stage process leveraging the multimodal capabilities of GPT-5. The pipeline is structured as follows:

\noindent \textbf{Stage 1: Initial Candidate Generation.} 
We present the original image $I$ and the instruction $T$ to the model. As shown in Figure~\ref{fig:prompts} (Prompt 1), the model is prompted to identify the body parts (e.g., hands) involved in the original action and propose a candidate verb $v'$ associated with a different body part. For instance, a hand-related action such as ``catch'' may be replaced with a foot-related one like ``kick''.

\noindent \textbf{Stage 2: Verification and Refinement.} 
To verify the candidate verb generated in Stage 1, we submit it to a second validation pass (refer to Prompt 2 in Figure~\ref{fig:prompts}). This stage confirms whether the new verb strictly targets a distinct body part and remains syntactically compatible with the original context. If the candidate fails this verification, the model automatically regenerates a more suitable replacement.

\noindent \textbf{Stage 3: Human Verification.} 
To ensure the quality and reliability of the dataset, we manually review all generated pairs. This manual screening filters out instances that exhibit semantic ambiguity or remain linguistically unnatural after the automated stages. Only high-quality, unambiguous samples are retained for the final evaluation set.

\begin{figure}[t] 
    \centering
    
    \begin{tcolorbox}[
        colback=gray!5, colframe=black, boxrule=0.5pt, arc=4pt,
        left=6pt, right=6pt, top=6pt, bottom=6pt,
        title=\textbf{Prompt 1 (Initial Generation)}
    ]
    You are a visual action analysis assistant. Given an image and a question that contains an action word associated with a specific body part, identify the body part implied by the original action and propose a replacement action word that involves a different body part.
    
    \vspace{0.5em}
    Output only the replacement action word (a single verb or a short verb phrase). Avoid actions that involve the same body part as the original.
    \end{tcolorbox}
    
    \vspace{0.5em} 
    
    \begin{tcolorbox}[
        colback=gray!5, colframe=black, boxrule=0.5pt, arc=4pt,
        left=6pt, right=6pt, top=6pt, bottom=6pt,
        title=\textbf{Prompt 2 (Verification and Refinement)}
    ]
    Verify the proposed replacement action word.
    
    (1) It must involve a body part different from the original action. \\
    (2) It must be semantically appropriate in the context of the question.
    
    \vspace{0.5em}
    If it satisfies both conditions, output it directly. Otherwise, output a new replacement action word that satisfies the rules. Output only the final action word.
    
    \vspace{0.5em} 
    \hrule
    \vspace{0.5em}
    
    Original question: \{question\_text\} \\
    Initial answer: \{initial\_answer\}
    \end{tcolorbox}
    
    \caption{The prompts used in our two-stage generation pipeline.} 
    \label{fig:prompts}
\end{figure}

\subsection{Comprehensive Examples}
\begin{table}[h]
    \centering
    \renewcommand{\arraystretch}{1.2}
    \resizebox{\linewidth}{!}{
    \begin{tabular}{l|c|c}
        \toprule
        \textbf{Original Question} & \textbf{Original Verb} & \textbf{Contrastive Verb} \\
        \midrule
        Does a woman \textbf{hold} bread in the image? & hold & eat \\
        Does a man \textbf{point} at food in the image? & point & look \\
        Does a man \textbf{ride} a skateboard in the image? & ride & hold \\ 
        Does a player \textbf{catch} a ball in the image? & catch & kick \\
        \bottomrule
    \end{tabular}
    }
    \caption{Action-contrastive samples generated by our pipeline.}
    \label{tab:generated_examples}
\end{table}

Table~\ref{tab:generated_examples} presents diverse examples of the generated contrastive samples from our dataset. As observed in the table, the original action verbs are replaced with plausible alternatives that involve different body parts (e.g., swapping the hand-related action ``catch'' with the foot-related action ``kick''). This ensures semantic coherence in the generated text while maintaining a clear visual distinction between the image regions relevant to the contrastive verb and the original verb.

\section{Evaluation Benchmarks and Setup}
\label{sec:appendix_benchmarks}
In this section, we provide detailed descriptions of the four benchmarks used to evaluate our RVE method. For all evaluations, we employ greedy decoding and fix the random seed to 55 for all experiments to ensure the reproducibility of the results.

\noindent \textbf{MMRel~\cite{nie2024mmrel}} is a large-scale relation-understanding benchmark that encompasses three core relation categories (spatial, action, and comparative), supporting both discriminative and generative evaluations. It comprises approximately 22.5K question-answer pairs, consisting of around 15K discriminative Yes/No pairs and 7.5K open-ended questions. The benchmark features diverse image sources, including real-world images as well as synthetic domains generated by SDXL and DALL-E. In our experiments, we evaluate the RVE method on the real-image, DALL-E, and open-ended subsets.

\noindent \textbf{R-Bench~\cite{wu2024evaluating}} is a benchmark designed to evaluate relationship hallucinations in LVLMs. Constructed on the nocaps validation set, it utilizes a Yes/No verification format to specifically target inter-object relationships. R-Bench comprises two complementary subsets: image-level questions, which assess the existence of relationships within the global scene, and instance-level questions, which evaluate local visual understanding by explicitly grounding subjects and objects with colored bounding boxes or masks. The benchmark contains 11,651 curated questions in total, consisting of 7,883 image-level and 3,768 instance-level samples. For instance-level evaluation, models are prompted with specific templates (e.g., ``\textit{Is there \{subject\} in the red box...}'') to verify precise relationship grounding.

\noindent \textbf{AMBER~\cite{wang2023amber}} is an LLM-free, multi-dimensional benchmark designed to systematically evaluate hallucinations across object existence, attributes, and relations. In this work, we specifically focus on the relation subset. Leveraging AMBER's high-quality human annotations, we adopt a discriminative QA format to induce relationship judgments using the specific prompt: ``\textit{Is there direct contact between the \{object 1\} and \{object 2\} in this image?}''. We regard such contact-based relations as essential action primitives, as they represent the physical interaction between objects (e.g., \textit{touching}). 

\noindent \textbf{POPE~\cite{li2023evaluating}} is a widely adopted benchmark for assessing object hallucinations in LVLMs. It evaluates models using Yes/No questions regarding object existence (e.g., ``\textit{Is there a \{object\} in the image?}''). The benchmark distinguishes three subsets based on the negative sampling strategy: Random, Popular, and Adversarial. Specifically, Random samples negative objects uniformly from the candidate set; Popular selects objects with the highest dataset frequency; and Adversarial targets objects that co-occur frequently with the ground-truth objects in the image. In this work, we conduct experiments specifically on the MSCOCO subset. This dataset comprises 3,000 test instances, consisting of 500 images with 6 questions per image.

\begin{figure}[h!]
    \centering
    \begin{tcolorbox}[
        colback=gray!5, 
        colframe=black, 
        boxrule=0.5pt, 
        arc=4pt,
        left=6pt, right=6pt, top=6pt, bottom=6pt, 
        title=\textbf{Evaluation Prompt}
    ]
    We would like to request your feedback on the performance of an AI assistant in response to the user question displayed above. The user asks a question regarding the action relationship between two objects.

    \vspace{0.5em}
    For your reference, the AI assistant is asked to answer with one sentence, which contains both objects and the action relationship.

    \vspace{0.5em}
    Please rate the response from the AI assistant based on the ground-truth answer. Each response receives an overall score on a scale of 0 to 10, where a higher score indicates the AI response is more consistent with the ground-truth answer.

    \vspace{0.5em}
    Please output the score for the Assistant.

    \vspace{0.5em} 
    \hrule
    \vspace{0.5em}

    \textbf{Ground-truth answer:} \{ground\_truth\} \\
    \textbf{AI assistant response:} \{generated\_answer\}
    \end{tcolorbox}
    \caption{The evaluation prompt used by the GPT-5-mini judge to evaluate open-ended generation tasks.}
    \label{fig:eval_prompt}
\end{figure}

\section{Details on the GPT-5-mini Evaluation}
\label{sec:appendix_eval_details}

To comprehensively evaluate performance on open-ended tasks, we employ GPT-5-mini as an automated judge. The model scores each response by comparing it against the ground-truth answer. The specific prompt used for this automated evaluation is presented in Figure~\ref{fig:eval_prompt}. The evaluator is instructed to focus specifically on the action relationship and the objects involved, assigning a score from 0 to 10.

\section{Additional Ablation Results}
\label{sec:appendix_ablation}

\begin{table}[h]
    \centering
    \renewcommand{\arraystretch}{0.9}
    \setlength{\tabcolsep}{3.5pt}
    
    \resizebox{\linewidth}{!}{
    \begin{tabular}{ccc|cc|cc}
        \toprule
        \multicolumn{3}{c|}{\textbf{Strategy}} & \multicolumn{2}{c|}{\textbf{LLaVA-NeXT-7B}} & \multicolumn{2}{c}{\textbf{ShareGPT4V-7B}} \\
        
        \textbf{Global} & \textbf{Sens.} & \textbf{Denoise} & \textbf{Acc} & \textbf{F1} & \textbf{Acc} & \textbf{F1} \\
        \midrule
        
        & & & 70.58 & 77.21 & 72.96 & 78.44 \\
        
        \midrule
        
        & \checkmark & & 73.02 & 78.58 & 73.43 & 78.65 \\
        
        \checkmark & & & 74.38 & 79.38 & 75.88 & 80.14 \\
        
        \midrule
        
        & \checkmark & \checkmark & 76.09 & 80.50 & 73.67 & 78.79 \\

        \checkmark & & \checkmark & \textbf{77.39} & \textbf{81.23} & \textbf{76.44} & \textbf{80.47} \\
        
        \bottomrule
    \end{tabular}
    }
    \caption{Additional ablation study on LLaVA-NeXT-7B and ShareGPT4V-7B regarding enhancement scopes and the denoising mask.}
    \label{tab:appendix_ablation}
    \vspace{-4mm}
\end{table}

To further demonstrate the generalizability of our method, we provide the ablation results on LLaVA-NeXT-7B and ShareGPT4V-7B in Table~\ref{tab:appendix_ablation}. Consistent with the findings in the main text, the results confirm that: (1) Global enhancement outperforms the sensitive-only approach by effectively aligning non-sensitive heads; and (2) the denoising mask further boosts performance by avoiding the enhancement of irrelevant regions. Ultimately, the combination of these two components achieves the best results across different LVLMs.

\section{Case Studies}
\label{sec:appendix_case_studies}
In this section, we present several case studies to evaluate model performance across multiple benchmark subsets. Specifically, we select examples from these three benchmarks, comprising seven diverse subsets: the real-image, DALL-E-generated, and open-ended subsets of MMRel, the image-level and instance-level (Mask and Box) tasks of R-Bench, and the relation subset of AMBER. In these examples, we can observe that our RVE method maintains effectiveness across diverse subsets compared to the vanilla baseline.

\begin{figure*}[t]
    \centering
    \includegraphics[width=1.0\linewidth]{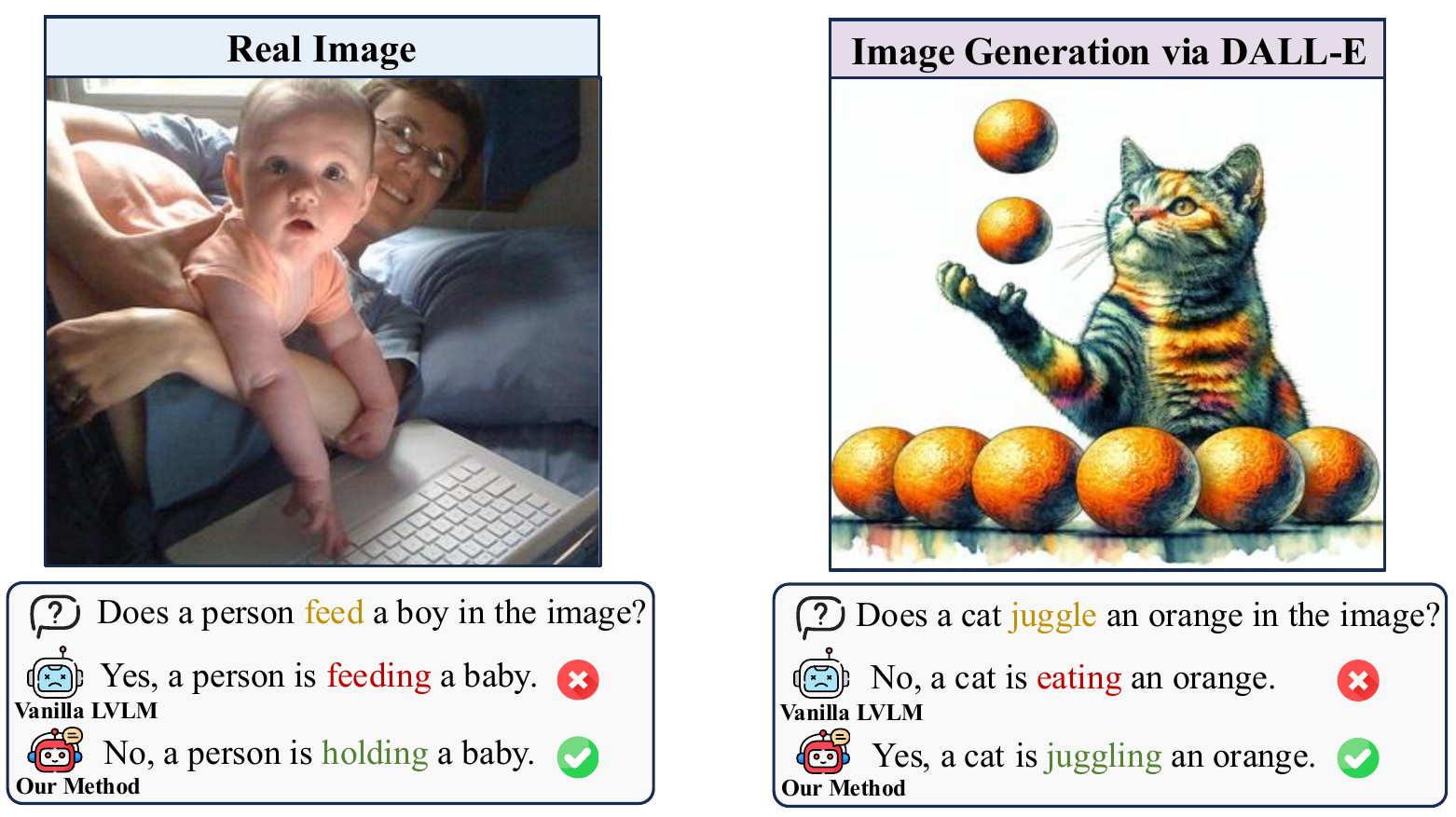}
    \caption{A case study of action-relation hallucinations on the MMRel benchmark across both real-world and DALL-E-generated scenarios.}
    \label{fig:case_studies}
\end{figure*}

\begin{figure*}[t]
    \centering
    \includegraphics[width=1.0\linewidth]{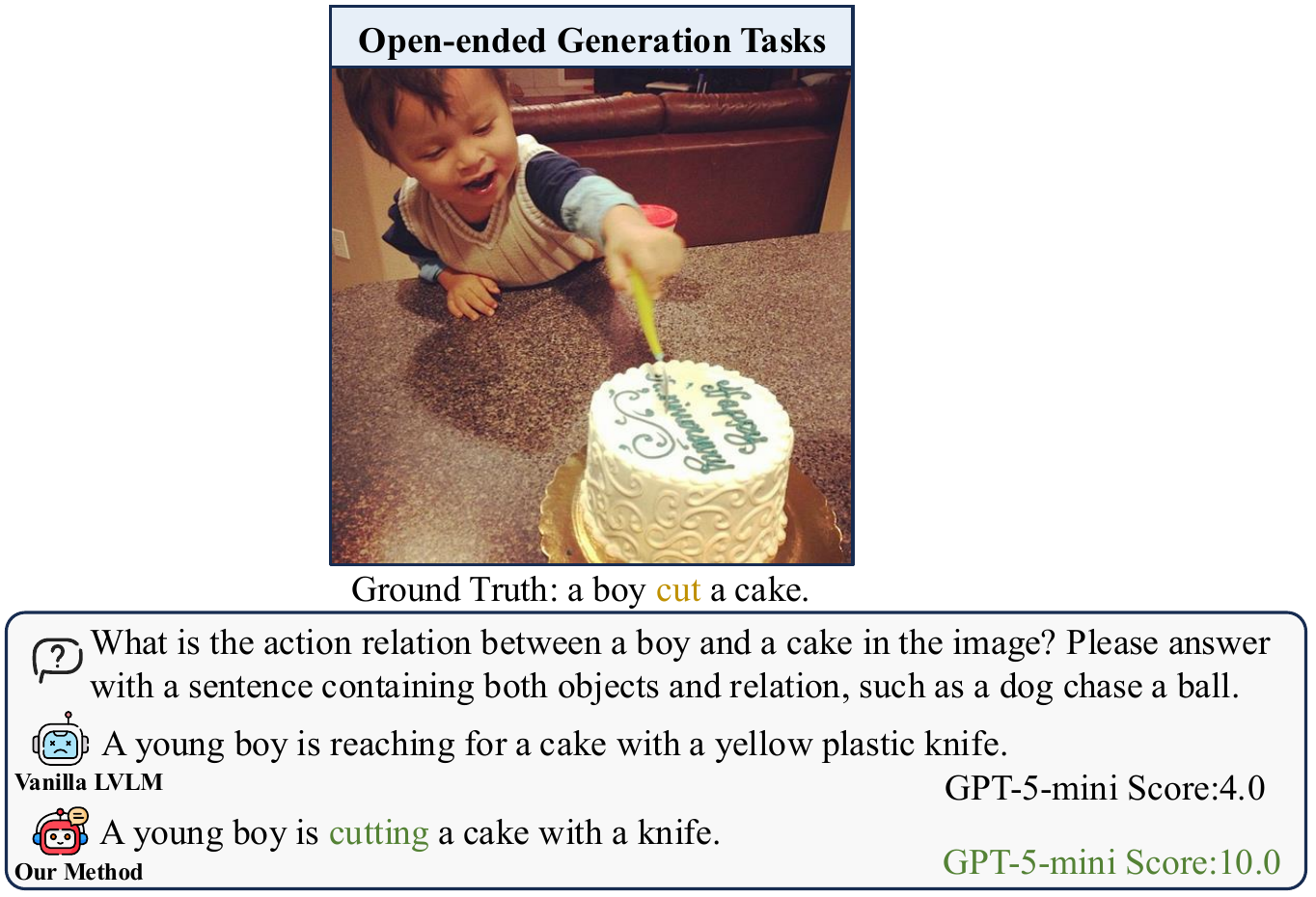}
    \caption{A case study of action-relation hallucinations in open-ended generation tasks on the MMRel benchmark.}
    \label{fig:case_open_ended}
\end{figure*}

\begin{figure*}[t]
    \centering
    \includegraphics[width=1.0\linewidth]{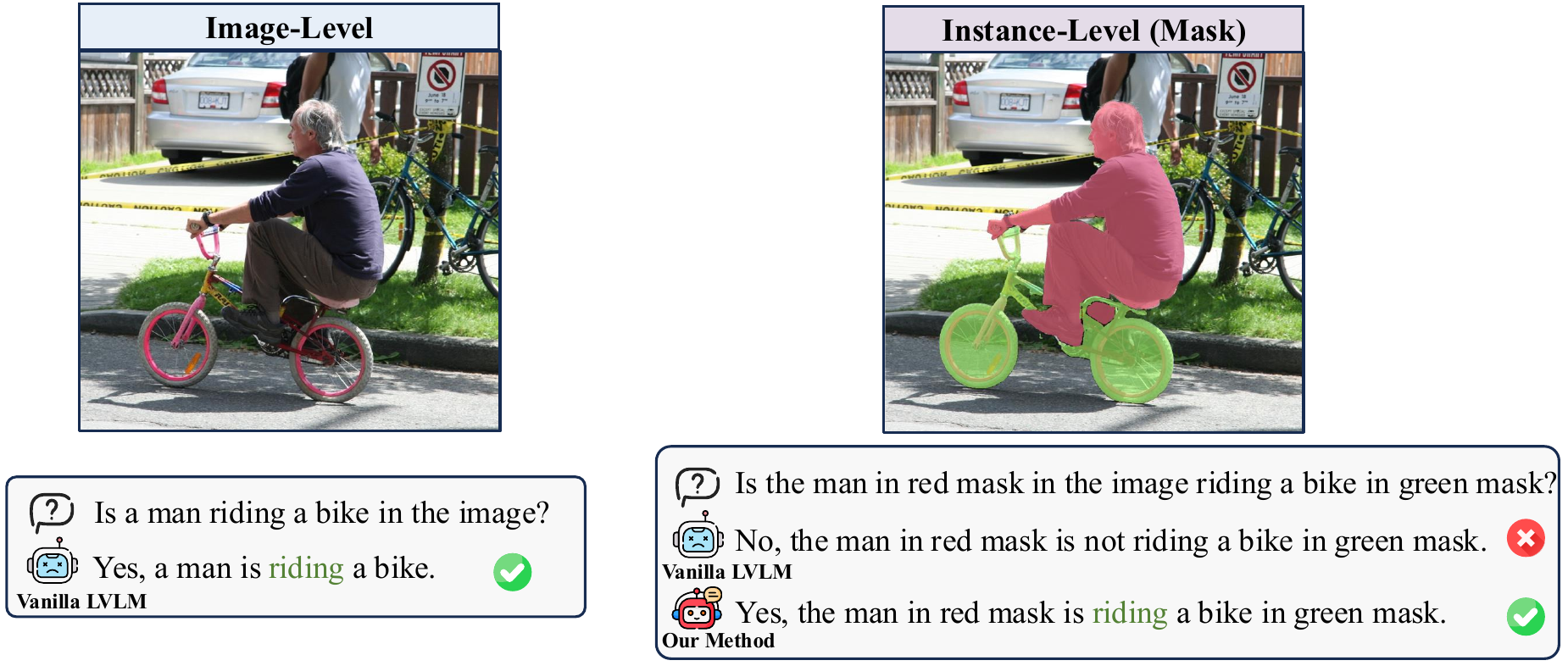}
    \caption{A case study comparing action-relation hallucinations between image-level and instance-level (Mask) tasks on the R-Bench benchmark.}
    \label{fig:case_rbench_mask}
\end{figure*}

\begin{figure*}[t]
    \centering
    \includegraphics[width=1.0\linewidth]{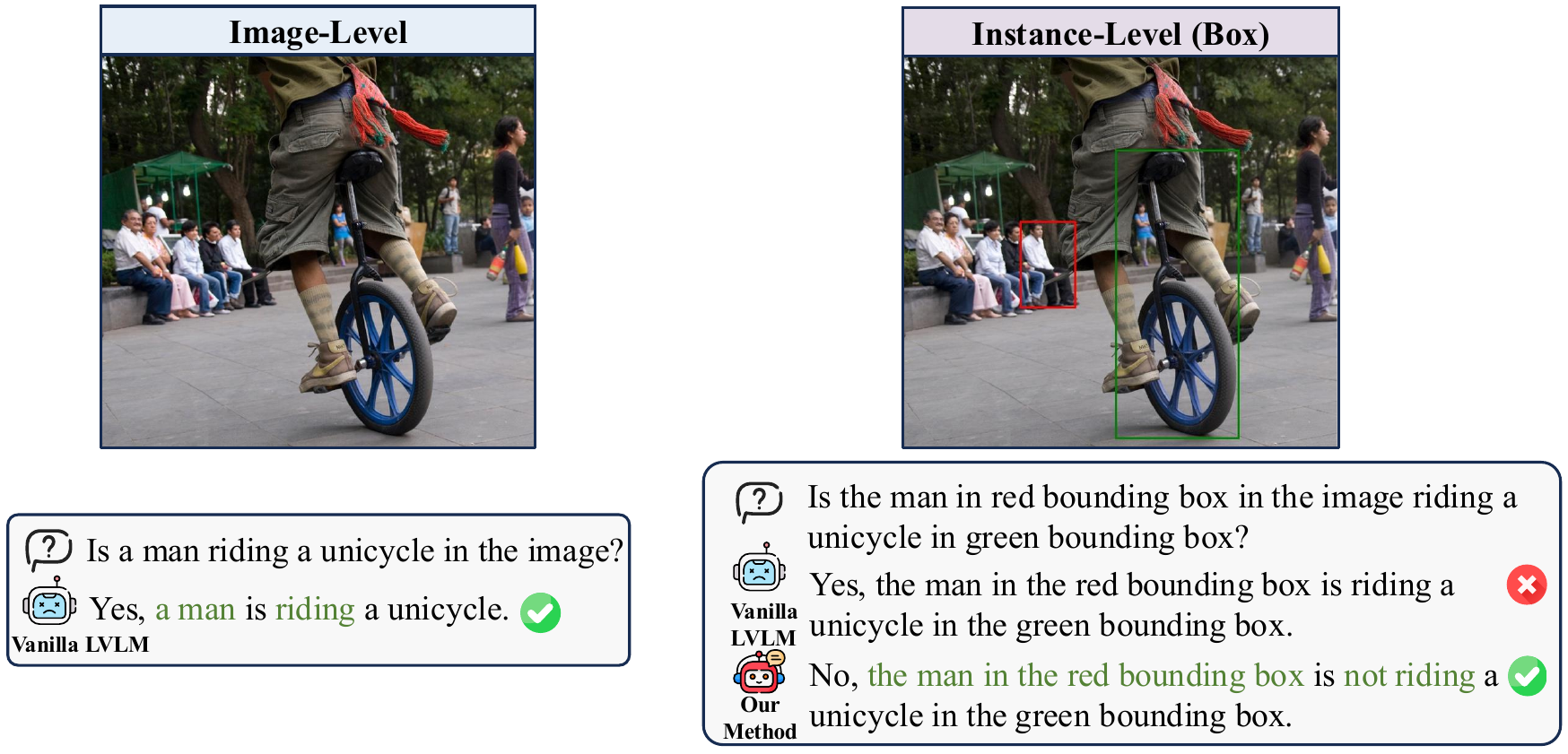}
    \caption{A case study comparing action-relation hallucinations between image-level and instance-level (Box) tasks on the R-Bench benchmark.}
    \label{fig:case_rbench_box}
\end{figure*}

\begin{figure*}[t]
    \centering
    \includegraphics[width=1.0\linewidth]{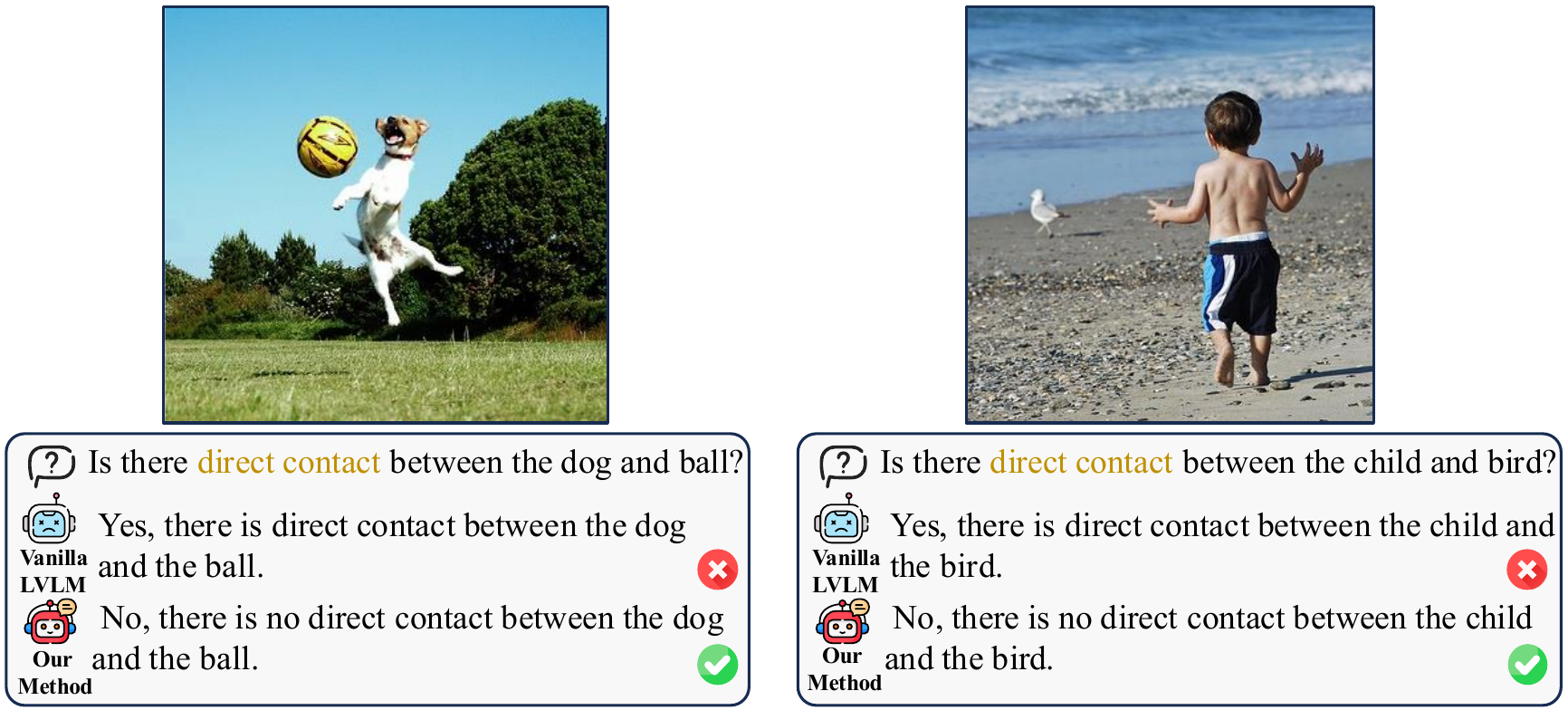}
    \caption{A case study of action-relation hallucinations on the AMBER benchmark.}
    \label{fig:case_amber}
\end{figure*}

\end{document}